\providecommand{\includeOpenFCAppendix}{1}
\documentclass[letterpaper]{article} 
\usepackage{aaai2027}

\usepackage[hyphens]{url}  
\usepackage{graphicx} 
\usepackage{natbib}  
\usepackage{caption} 
\usepackage{booktabs}
\usepackage{colortbl}
\usepackage{amsmath}
\usepackage{multirow}
\usepackage{listings}
\lstdefinestyle{openfcprompt}{
  basicstyle=\footnotesize\ttfamily,
  numbers=none,
  showstringspaces=false,
  keepspaces=true,
  columns=fixed,
  basewidth=0.55em,
  breaklines=true,
  breakatwhitespace=true,
  breakindent=0pt,
  frame=tb,
  framerule=0.4pt,
  framesep=3pt,
  xleftmargin=0pt,
  xrightmargin=0pt,
  aboveskip=4pt,
  belowskip=6pt,
  captionpos=t
}
\title{OpenFC: Learning Verification Policies towards Open-Search Fact Checking}

\author{
    Xinming Wang\textsuperscript{\rm 1},
    Kaixiang Qiu\textsuperscript{\rm 1},
    Yansong Lin\textsuperscript{\rm 2},
    Chunji Lv\textsuperscript{\rm 3},\\
    Yi Chen\textsuperscript{\rm 1},
    Boran Wang\textsuperscript{\rm 4},
    Hong-Ming Yang\textsuperscript{\rm 1},
    Xu-Yao Zhang\textsuperscript{\rm 1}\thanks{Corresponding author.}
}

\affiliations{
    \textsuperscript{\rm 1}Institute of Automation, Chinese Academy of Sciences\\
    \textsuperscript{\rm 2}University of Electronic Science and Technology of China\\
    \textsuperscript{\rm 3}Beijing Institute of Technology ~~
    \textsuperscript{\rm 4}Nankai University\\
    \{wangxinming2024, qiukaixaing2025\}@ia.ac.cn, xyz@nlpr.ia.ac.cn
}

\begin{document}

\maketitle

\begin{abstract}
Open-search fact checking is not merely retrieval followed by classification, but a sequential decision problem in which every query, source visit, and stopping decision reshapes the evidence available for verification. Yet existing systems often distribute these decisions across predefined pipelines or separately prompted modules rather than learning them as a unified task-specific policy. We introduce \textbf{OpenFC}, a unified verification-policy training framework that post-trains Qwen3-8B as a compact next-action controller over reasoning, evidence acquisition, and stopping.
OpenFC learns this policy in two stages. \textbf{Stepwise-Calibrated Cold Start (SCCS)} uses a strong training-time supervisor to review post-initial reasoning, tool-use, and stopping proposals before execution, producing reliable trajectories for supervised fine-tuning without access to gold verdicts. \textbf{Verification-Aware Reinforcement Learning (VA-RL)} then improves the cold-start policy on unresolved claims through budget-aware tool rewards, label-aware advantage reweighting, and localized response masking.
Across six fact-checking benchmarks, OpenFC achieves 70.39\% average accuracy and 63.30\% macro-F1, the highest overall averages among the evaluated methods. Stage-wise ablations further show that SCCS and VA-RL provide complementary gains, supporting the design of the two-stage training framework. These results position OpenFC as a strong and effective framework for open-search fact-checking. We will open-source our code and release the model checkpoints to support reproducibility.
\end{abstract}


\section{Introduction}

Open-search fact checking is not merely label prediction after retrieval; it is sequential control over which evidence becomes available.
Real-world benchmarks require evidence to be gathered from the open web rather than supplied in advance~\citep{schlichtkrull2023averitec,chen2024complex}.
Raw web retrieval further introduces source reliability, temporal validity, and evidence coverage constraints that a verifier must manage throughout the interaction~\citep{schlichtkrull2024generating,sriram2024contrastive}.
At each step, the verifier must decide whether to reason over the current evidence, issue a new query, inspect a source, or stop with a verdict~\citep{yao2023react,xie2025fire}.
We call the decision rule that maps the claim, interaction history, observed evidence, and remaining budget to one of these next actions a \emph{verification policy}.
Unlike a generic search policy, a verification policy must seek both supporting and disconfirming evidence~\citep{venktesh2025factir}, distinguish conflicting evidence from insufficient evidence~\citep{schlichtkrull2023averitec,glockner2024ambifc}, and abstain when the available record does not justify a definitive label~\citep{thorne2018fever}.

\begin{figure}[t]
\centering
\includegraphics[width=0.98\columnwidth]{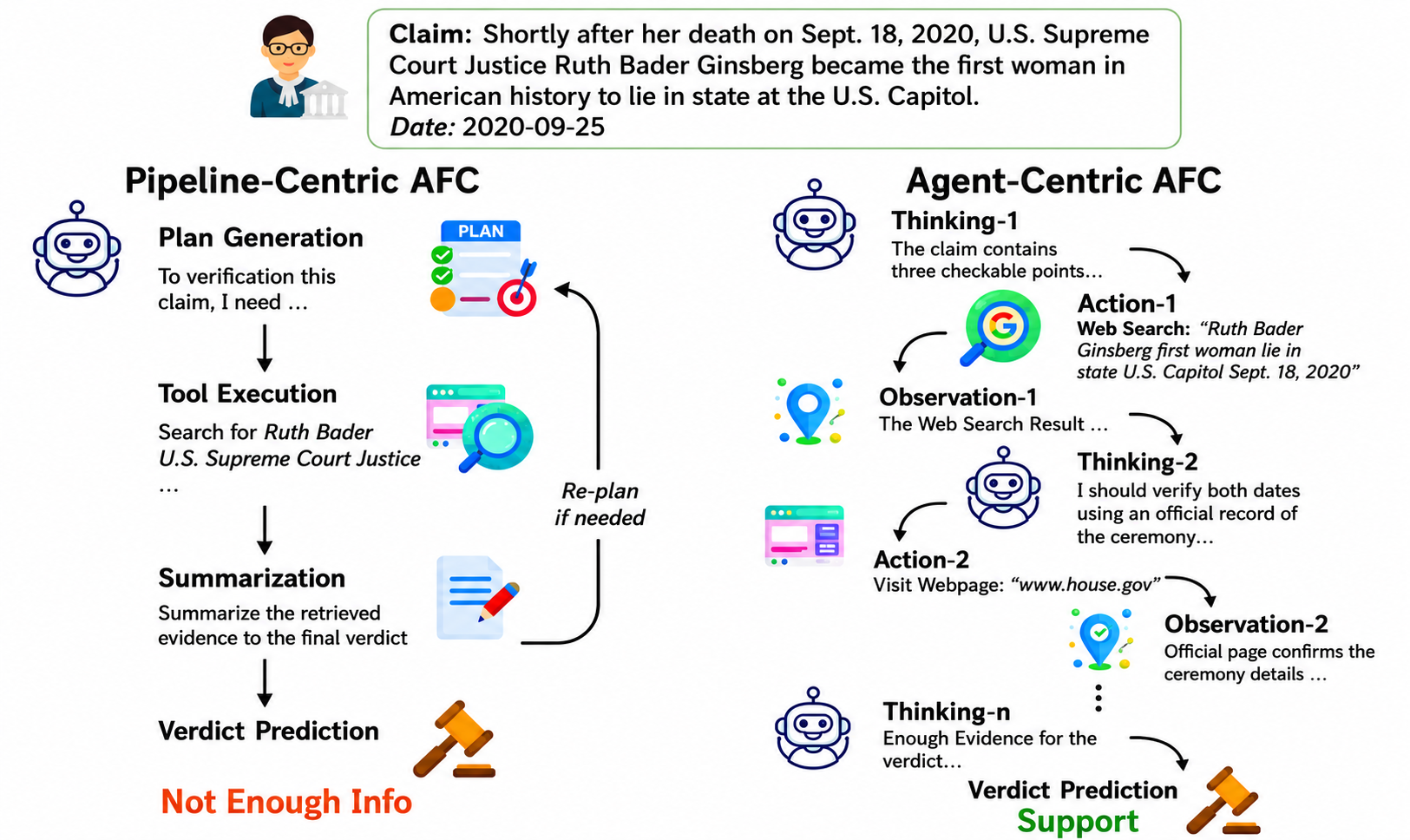}
\caption{Pipeline-centric automatic fact-checking (AFC) assigns verification to staged modules, whereas agentic AFC integrates reasoning, retrieval, and stopping in a ReAct-style.}
\label{fig:pipeline-vs-agent-afc}
\vspace{-6pt}
\end{figure}

Existing automatic fact-checking (AFC) systems typically encode much of the verification policy in a predefined pipeline, assigning retrieval, evidence synthesis, and verdict prediction to explicit stages~\citep{thorne2018fever,wadden2020scifact}.
Recent systems make this pipeline more adaptive: ClaimCheck combines compact language-model components with web evidence~\citep{putta2025claimcheck}, DEFAME dynamically orchestrates tools and evidence sources~\citep{braun2025defame}.
Nevertheless, as shown in Figure~\ref{fig:pipeline-vs-agent-afc}, key control decisions in pipeline-centric systems remain specified through prompted or separately trained modules, external orchestration, or hand-designed rules, rather than being jointly internalized by a task-trained verification policy.
In parallel, general-purpose search agents follow the ReAct~\cite{yao2023react} paradigm and learn multi-turn tool-use policies for question answering and research synthesis~\citep{jin2025searchr1,tongyi2025deepresearch}, but their broad information-seeking objectives do not explicitly teach the fact-checking-specific behavior defined above.
The underexplored question is therefore not whether retrieval can be iterative, but whether joint next-action control for open-search fact checking can be learned by one compact and task-specific controller rather than specified through prompted or separate modules.

To this end, we introduce \textbf{OpenFC}, a task-specific unified verification-policy framework that post-trains Qwen3-8B~\citep{yang2025qwen3} as the action policy in a fixed verification environment. It is task-specific because its trajectory supervision, verdict-aware optimization, and stopping behavior are tailored to evidentiary adjudication rather than generic information seeking; it is unified because a single learned controller selects reasoning continuations, evidence-acquisition actions, and termination decisions. At inference, OpenFC runs a ReAct-style loop: conditioned on the claim, evidence cutoff, and accumulated observations, it updates its verification state, calls \textsc{Search} or \textsc{Visit} to gather evidence, and terminates with an evidence-grounded verdict once information is sufficient for adjudication.
We train the policy in two stages. \textbf{Stepwise-Calibrated Cold Start (SCCS)} uses a strong supervisor to review each post-initial reasoning step, tool call, or terminal decision before execution, retaining or revising it based on the observed evidence. Correct calibrated trajectories are then used for supervised fine-tuning. \textbf{Verification-Aware Reinforcement Learning (VA-RL)} further trains the cold-start policy on unresolved claims. Building on GRPO, it combines a tool budget reward~\citep{xie2026slimsearcher}, label-aware advantage reweighting, and token-level response masking. The tool reward is applied only when a correct rollout provides a valid efficiency reference, while masking removes gradients from external execution failures but preserves penalties for policy-generated malformed calls. Together, these mechanisms promote accurate, valid, and efficient verification.

We evaluate OpenFC on benchmarks spanning open-domain, scientific, recent, and numerical-temporal fact checking. OpenFC achieves 70.39\% accuracy and 63.30\% macro-F1 and attains the highest performance among other methods. Comparisons among shared-interface agents hold the prompt, tools, and interaction budget fixed; native-search and separately orchestrated systems are included as end-to-end references rather than policy-only controls.

Our contributions are summarized as follows:
\begin{itemize}
\item We formulate open-search fact checking as learning a unified next-action verification policy within a fixed environment.
\item We propose a task-specific unified verification-policy training framework that combines stepwise-calibrated cold start with verification-aware reinforcement learning, internalizing reasoning, evidence acquisition, and stopping in one compact controller.
\item We report gains across diverse fact-checking benchmarks, with stage-wise ablations examining the contribution of the proposed training components.

\end{itemize}

\section{Related Work}

\subsection{Automated Fact Checking}

Automated fact checking (AFC) is commonly organized around claim analysis, evidence retrieval, evidence assessment, and verdict prediction~\citep{guo2021survey,vykopal2024survey}.
Benchmarks span closed-corpus verification~\citep{thorne2018fever,jiang2020hover,wadden2020scifact}, multi-domain claims~\citep{augenstein2019multifc}, and evidence retrieval over the open web~\citep{schlichtkrull2023averitec,chen2024complex}.
LLM systems make this pipeline more adaptive through executable decomposition and modular tool use~\citep{pan2023programfc,chern2023factool,li2024selfchecker,xmwang}; \textsc{DEFAME} selects among tools and evidence sources, while \textsc{FIRE} couples iterative retrieval with confidence-based stopping~\citep{braun2025defame,xie2025fire}.
In these representative designs, however, control remains distributed across prompted modules, external orchestration, or explicit decision criteria.
OpenFC's distinction is therefore not iterative retrieval, but post-training one compact controller to jointly decide how to reason, reformulate queries, inspect sources, and stop.

\subsection{Open Search and Deep Search}

Open-search agents formulate information seeking as a sequential interaction.
\textsc{WebGPT} and \textsc{ReAct} established browser-assisted generation and interleaved reasoning and acting~\citep{nakano2021webgpt,yao2023react}, while subsequent methods moved retrieval timing and critique into the reasoning loop~\citep{trivedi2022ircot,jiang2023flare,asai2023selfrag,jeong2024adaptiverag}.
More recent systems directly post-train multi-turn search policies with supervised trajectories or outcome-based reinforcement learning~\citep{lv2026pcsd,li2025searcho1,jin2025searchr1,song2025r1searcher,chen2025research,zheng2025deepresearcher,wu2025webdancer,li2025websailor,tongyi2025deepresearch}.
This literature establishes that search depth, query reformulation, and tool use can be learned.
Its supervision and rewards, however, primarily target question-answer accuracy or report quality rather than the evidence states required for adjudication.
Open-search fact checking additionally requires seeking supporting and disconfirming evidence, distinguishing evidentiary conflict from insufficiency, and stopping only when the record justifies a verdict.
OpenFC uses the same interaction paradigm, distills adjudication-specific behaviors via stepwise-calibrated SFT, and then refines the policy with outcome rewards, per-rollout format rewards, and a bounded group-conditional tool reward for search efficiency. 


\begin{figure*}[t]
\centering
\includegraphics[width=\textwidth]{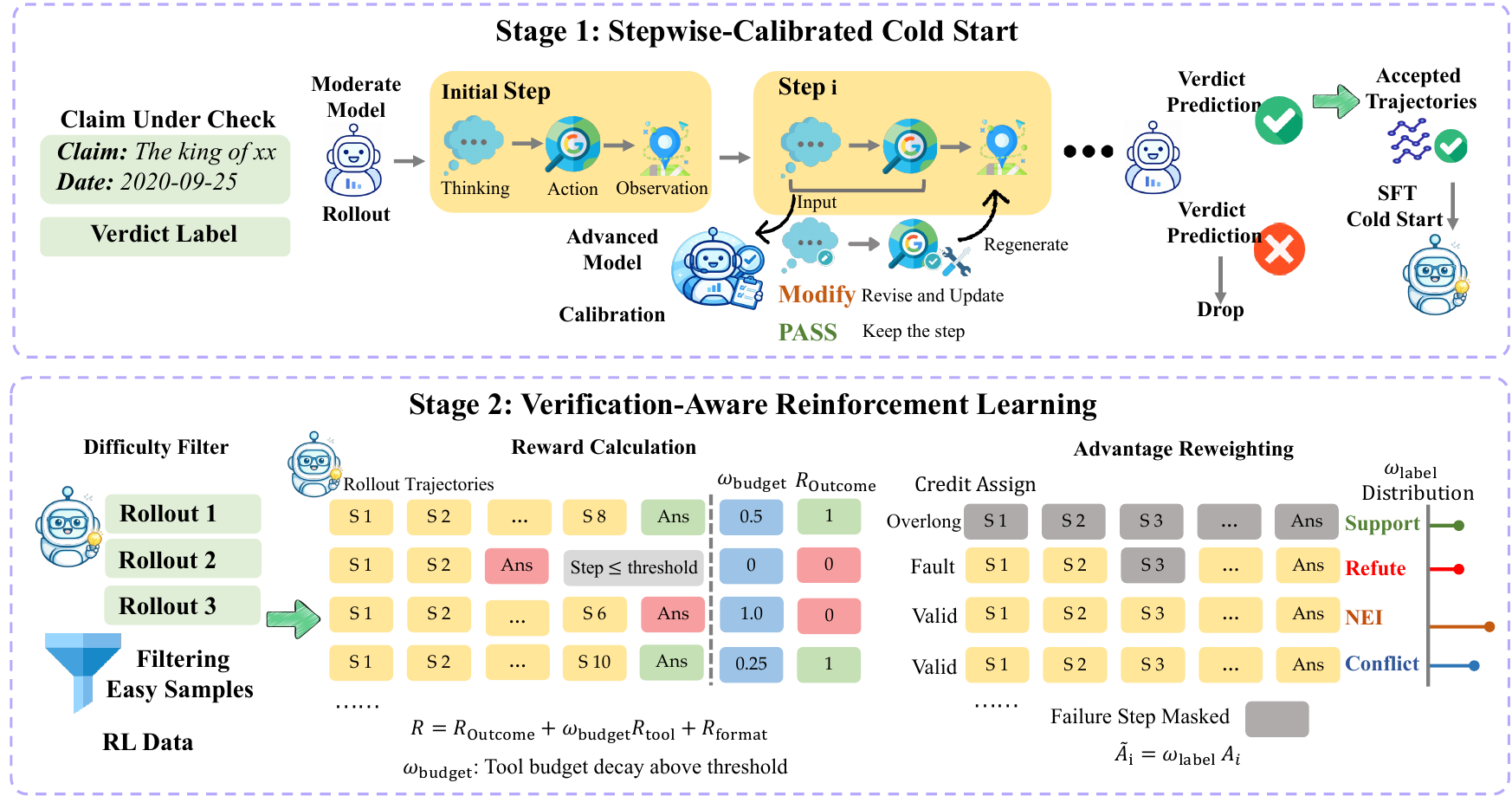}
\caption{OpenFC training process. Stage 1 uses Stepwise-Calibrated Cold Start (SCCS) to calibrate policy rollouts before execution; Stage 2 applies Verification-Aware Reinforcement Learning (VA-RL) to improve fact-checking generalization.}
\label{fig:openfc-method-overview}
\end{figure*}

\section{Method}

\subsection{Problem Formulation}

\subsubsection{Interaction Model}
Given a claim $x$ and evidence cutoff date $d$, let $c=(x,d)$. OpenFC models fact checking as a verification policy $\pi_\theta$ over reasoning, evidence seeking, and stopping, with at most $M$ tool calls. At turn $t$, the policy conditions on
\begin{equation}
h_t=(c,s_1,o_1,\ldots,s_{t-1},o_{t-1})
\end{equation}
and generates $s_t=(r_t,a_t)\sim\pi_\theta(\cdot\mid h_t)$, where $r_t$ is a reasoning continuation and $a_t$ is a tool or terminal action.
For \textsc{Search} or \textsc{Visit}, the environment returns $o_t=\mathcal E(a_t)$ and appends $(s_t,o_t)$ to form $h_{t+1}$.
A terminal action ends the rollout with verdict $\hat y$ and a concise evidence-based rationale.

The trajectory is
\begin{equation}
\tau=(c,s_1,o_1,\ldots,s_{T-1},o_{T-1},s_T),
\end{equation}
where $s_T=(r_T,a_T)$ is terminal and $T-1\leq M$.
Thus, a single policy chooses the next reasoning and action, while observations are environment-generated context.

The verdict $\hat y$ is one of \textsc{Supported}, \textsc{Refuted}, \textsc{Conflicting Evidence}, or
\textsc{NEI}, where \textsc{NEI} denotes \textsc{Not Enough Information}.
\textsc{Conflicting Evidence} means credible, directly relevant evidence both supports and contradicts the claim, with no sufficient basis to resolve the conflict.
\textsc{NEI} means the retrieved evidence is insufficient for any other verdict; it does not imply the claim is false. Although label definitions vary across datasets, they generally fall into these four categories, so we adopt them as a unified standard.

\subsubsection{Environment Settings}
OpenFC uses two tools following common deep-search interfaces~\citep{tongyi2025deepresearch}.
\textsc{Search}$(q_t)$ submits a query through the Serper API\footnote{\url{https://serper.dev/}} and returns ranked results with snippets and URLs.
\textsc{Visit}$(u_t,g_t)$ fetches a selected URL through Jina Reader\footnote{\url{https://jina.ai/reader/}} and asks \texttt{gpt-oss-20b} to produce a goal-conditioned page summary.
All observations are recorded in the transcript by the environment.
Following realistic fact-checking settings~\citep{schlichtkrull2023averitec,chen2024complex,braun2025defame}, each instance is associated with a fixed cutoff date $d$, defined as the release date of the claim.
The environment enforces this temporal boundary at both interfaces: \textsc{Search} removes results published after $d$ or without a resolvable publication date, while \textsc{Visit} rejects URLs whose publication date is later than $d$ or cannot be resolved.
We also block professional fact-checking domains and URL-path patterns from both \textsc{Search} results and \textsc{Visit} requests for all shared-interface methods. Each such method uses the same fact-checking prompt, identical \textsc{Search} and \textsc{Visit} implementations, and a total budget of 30 tool calls. If a \textsc{Search} action lists multiple queries, only the first three are run; each \textsc{Visit} action dispatches a single URL, so pages are not fetched in parallel. Together, the temporal filter and fact-check blocklist exclude undated or post-cutoff pages and direct professional verdicts. The technical supplement specifies the domain-blacklisting rules.

\subsection{Stepwise-Calibrated Cold Start (SCCS)}

The cold-start stage instills a verification policy for fact-checking via multi-step tool-interaction trajectories. To avoid the high cost of fully proprietary rollouts, we use Tongyi-DeepResearch-30B-A3B as a mid-tier policy model and GPT-5.1 as a stronger stepwise supervisor. At each step, the calibration model reviews the policy model’s reasoning and tool call, keeping it or revising it as needed. This iterative calibration guides the policy model through the rollout and yields high-quality trajectories for later training.

\subsubsection{Seeded Rollout}

We draw seed claims from the training splits of AVeriTeC, QuanTemp++, and SciFact~\citep{schlichtkrull2023averitec,venktesh2025quantempplus,wadden2020scifact}.
Each instance is $(c_i,y_i)$, where $y_i$ is the gold verdict and $c_i = (x_i, d_i)$ is the claim with evidence cutoff date.
Starting from $c_i=(x_i,d_i)$, the rollout policy proposes reasoning $r_t$ from the calibrated history and then a \textsc{Search}, \textsc{Visit}, or terminal action $a_t$.
The policy model stops at the boundary where the environment must provide the tool response, preventing the policy model from inventing an observation.

\subsubsection{Calibration Intervention}
Rather than generating the complete trajectory itself, GPT-5.1 intervenes locally at each transition. It reviews the rollout policy's proposed reasoning and action before execution, assessing whether the proposal follows from the evidence observed so far, whether the tool call is valid and purposeful, and whether a terminal decision is sufficiently supported.
Calibration begins at the second step and the initial rollout bypasses the supervisor, with $s_1=\widetilde s_1=(r_1,a_1)$; a tool action yields $o_1=\mathcal E(a_1)$, whereas a terminal action ends the rollout.
This unsupervised first action anchors calibration in an actual environment response rather than a supervisor-generated premise.
For each $t\geq2$, the rollout verification policy proposes $\widetilde s_t=(\widetilde r_t,\widetilde a_t)$.
GPT-5.1 receives the claim and cutoff datetime $c_i$, the immediately preceding policy step $s_{t-1}=(r_{t-1},a_{t-1})$, its resulting observation $o_{t-1}$ proposed step $\widetilde s_t$, and step index $t$ and generate the calibrated decision $(\delta_t, s_t)$:
\begin{equation}
\begin{aligned}
(\delta_t,s_t)
&=\mathcal C_{\mathrm{ver}}
(c_i,s_{t-1}, o_{t-1},\widetilde s_t,t),\\
\delta_t&\in\{\textsc{Keep},\textsc{Modify}\}.
\end{aligned}
\end{equation}
Providing $(s_{t-1},o_{t-1})$ lets the supervisor judge whether the proposal is a coherent continuation of the latest interaction. Each calibration decision depends only on the local context, observed evidence, and the tool-use contract. The supervisor prompt is in the appendix.
Under $\textsc{Keep}$, the step executes unchanged with $s_t=\widetilde{s}_t$. Under $\textsc{Modify}$, the supervisor revises unsupported reasoning, malformed tool use, or an unjustified stopping decision at the transition boundary, before it affects the environment. For an unsupported terminal proposal, the supervisor revises the reasoning and requires the rollout policy to regenerate $a_t$ instead of supplying a verdict, preserving the rollout policy as the acting verification policy. Only the resulting full step $s_t=(r_t,a_t)$ is executed and added to the trajectory. A tool action yields $o_t=\mathcal{E}(a_t)$ and updates the history with $(s_t,o_t)$; a terminal action ends the rollout. Because the calibrated step determines both the next observation and subsequent policy context, each modification can change the downstream evidence path rather than merely edit an isolated response.

\subsubsection{Successful-Trajectory Selection}
Stepwise calibration improves individual transitions but does not ensure correct final verdicts. We therefore keep only trajectories whose final verdict matches the gold label and whose interaction structure is valid. A valid trajectory has legal tool schemas, no truncation, exactly one terminal verdict, and more than three tool calls. The >3-call requirement is a demonstration-selection heuristic favoring trajectories with a retrieval–inspection–refinement cycle, exposing the target policy to query formulation, source examination, and evidence-based revision. The resulting cold-start SFT corpus contains 4,361 trajectories; their detailed distribution appears in Table~\ref{tab:sft-statistics}. Each retained trajectory is serialized as one SFT example, with loss computed only on policy-generated reasoning, tool-action, and verdict tokens, while claims and environment observations are masked.

\begin{table}[t]
\centering
\footnotesize
\setlength{\tabcolsep}{4.5pt}
\begin{tabular}{@{}lrrrr@{}}
\toprule
\textbf{Label} & \textbf{AVeriTeC} & \textbf{QuanTemp++} & \textbf{SciFact} & \textbf{Total} \\
\midrule
Refuted & 1,087 & 752 & 158 & 1,997 \\
Supported & 536 & 631 & 324 & 1,491 \\
CE & 165 & 426 & -- & 591 \\
NEI & 282 & -- & -- & 282 \\
\midrule
\textbf{Total} & \textbf{2,070} & \textbf{1,809} & \textbf{482} & \textbf{4,361} \\
\bottomrule
\end{tabular}
\caption{Accepted claim-level trajectories for cold-start SFT.}
\label{tab:sft-statistics}
\end{table}

\subsection{Verification-Aware Reinforcement Learning (VA-RL)}
Building on the cold-start policy, we apply RL to unresolved claims. The objective combines verdict correctness with tool-budget rewards, while advantage reweighting on token-level masking and label distribution improves training stability and performance.

\subsubsection{Difficulty-Filtered Claims}
The cold start initializes the verification policy. RL then focuses on unresolved claims, using three independent cold-start rollouts to filter consistently solved examples and retain those that remain difficult. Concentrating training on these claims avoids spending the group-sampling budget on cases for which the cold-start policy already produces uniformly correct answers.
\begin{equation}
\widehat p_i
=
\frac{1}{3}
\sum_{j=1}^{3}
\mathbf 1[\widehat y_{ij}=y_i],
\qquad
\mathcal D_{\mathrm{RL}}
=
\{c_i:\widehat p_i<1\}.
\end{equation}
Claims solved in all three rollouts are removed.
The retained pool comprises partially solved claims, for which one or two of three rollouts are correct (Pass@3), and all-failed claims, for which none of the three rollouts is correct.
This filtering focuses the rollout budget on unresolved verification behavior.
After difficulty filtering and label balancing, the RL pool contains 7,483 training rows, including 4,794 partially solved and 2,689 all-failed rows.
Table~\ref{tab:rl-statistics} summarizes the resulting distribution.

\begin{table}[t]
\centering
\footnotesize
\setlength{\tabcolsep}{4.5pt}
\begin{tabular}{lrrrr}
\toprule
\textbf{Label} & \textbf{AVeriTeC} & \textbf{QuanTemp++} & \textbf{SciFact} & \textbf{Total} \\
\midrule
Refuted    & 1,162 & 2,409 & 180 & 3,751 \\
Supported  &   496 &   757 & 206 & 1,459 \\
CE         &   194 & 1,523 &   -- & 1,717 \\
NEI        &   556 &     -- &  -- &   556 \\
\midrule
\textbf{Total} & \textbf{2,408} & \textbf{4,689} & \textbf{386} & \textbf{7,483} \\
\midrule
Pass@3        & 1,587 & 2,894 & 313 & 4,794 \\
All Failure   &   821 & 1,795 &  73 & 2,689 \\
\bottomrule
\end{tabular}
\caption{Label distribution of the RL training data.}
\label{tab:rl-statistics}
\vspace{-6pt}
\end{table}

\subsubsection{Reward with Tool Budget}
Verdict-only supervision underdetermines the verification process: trajectories that reach the same answer receive identical outcome credit despite substantial differences in evidence acquisition and tool efficiency.
We therefore combine outcome, tool-budget, and format rewards.
Let $R_\mathrm{outcome}=\mathbf 1[\widehat y_i=y_i]$ indicate verdict correctness:
\begin{equation}
R_i
=
R_\mathrm{outcome}
+
\omega_{\mathrm{budget},i}R_{\mathrm{tool}}
+
R_{\mathrm{format},i}.
\end{equation}
The format reward is evaluated for every rollout, independently of verdict correctness: $R_{\mathrm{format},i}=0.05$ when all \verb|<think>| and \verb|<tool_call>| blocks are well formed and properly matched, and $0$ otherwise. The tool reward is bounded by $R_{\mathrm{tool}, i}\leq 0.15$; together with the outcome reward of at most $1$, the total reward is capped at $1.20$.

For the tool reward, we encourage efficient evidence acquisition without incentivizing premature termination~\cite{wang2025otc}. Let $t_i$ be the number of valid \textsc{Search} and \textsc{Visit} calls, and set the minimum-depth threshold to $\tau=3$, matching the calibration. This threshold is an exploration floor: trajectories with at most $\tau$ calls receive no tool bonus, so the policy cannot gain efficiency credit by stopping before a basic retrieval--inspection--refinement cycle.
For rollout group $g$, define $\mathcal C_g^+=\{j\in g:\widehat y_j=y_i,\ t_j>\tau\}$ as the correct rollouts that exceed this. If $\mathcal C_g^+=\emptyset$, no valid efficiency reference exists and the tool component is set to zero for all rollouts in $g$. Otherwise, set $t_{\min}=\min_{j\in\mathcal C_g^+}t_j$, giving a claim-specific efficiency reference, since different claims may require different levels of evidence acquisition. We compute
\begin{equation}
\omega_{\mathrm{budget},i}=
\begin{cases}
0, & \mathcal C_g^+=\emptyset\ \text{or}\ t_i\leq\tau,\\
\min\!\left\{1,\,2^{-\lfloor (t_i-t_{\min})/2\rfloor}\right\},
& \text{otherwise}.
\end{cases}
\end{equation}
When a correct-rollout reference exists, the coefficient is bounded by one and is halved for every two calls beyond $t_{\min}$. The stepwise decay tolerates small variations in valid search paths but progressively reduces credit for redundant retrieval or inspection. 


\subsubsection{Advantage Reweighting}
To enable finer-grained credit assignment, we combine token-level masking with label-aware advantage reweighting. Let $B_{i,k}$ mask policy-generated tokens, excluding prompts, tool observations, and padding. We apply an additional local mask only to environment-side failures not attributable to the policy. A tool-call span is eligible when the action is syntactically valid, passes validation, reaches execution, but its result is unusable due to an API timeout or network failure, or temporary webpage unavailability. We collect these spans in $\mathcal I_i^{\mathrm{env}}$, which is empty if no such failure occurs. We define
\begin{equation}
M_{i,k}
=
B_{i,k}\mathbf{1}\!\left[k\notin\mathcal I_i^{\mathrm{env}}\right].
\end{equation}
This response mask is used only in the token-level policy-gradient loss. It prevents low returns from external outages or corrupted backends from penalizing otherwise reasonable tool actions, while preserving gradients for policy-induced errors. The mask does not change the scalar rollout reward or the grouping used for advantage normalization. For groups with nonzero reward variance, we compute
\begin{equation}
A_i
=
\frac{R_i-\mu_g}
{\sigma_g},
\qquad
\widetilde A_i
=
\omega_{\mathrm{label}}(y_i^\star)A_i .
\end{equation}
Because all rollouts for a claim share the same gold verdict, $\omega_{\mathrm{label}}(y_i^\star)$ preserves their within-group ranking while controlling claim-level update magnitude. We set $\omega_{\mathrm{label}}(\textsc{Supported})=\omega_{\mathrm{label}}(\textsc{Refuted})=1.0$, $\omega_{\mathrm{label}}(\textsc{CE})=1.2$, and $\omega_{\mathrm{label}}(\textsc{NEI})=1.5$. The larger weights emphasize these harder cases, which require broader evidence, conflict resolution, and avoiding premature verdicts. As shown in Figure~\ref{fig:advantage-reweighting}, this reweighting produces more balanced label-level learning signals during training.

\section{Experiments}

\begin{table*}[t]
\centering
{\footnotesize
\renewcommand{\arraystretch}{1.10}
\setlength{\tabcolsep}{2pt}
\begin{tabular*}{\textwidth}{
  @{\extracolsep{\fill}}l*{14}{c}@{}
}
\toprule

\multicolumn{1}{c}{
  \multirow{3}{*}{\textbf{Method}}
}
& \multicolumn{6}{c}{\textbf{In-distribution Dataset}}
& \multicolumn{6}{c}{\textbf{Out-of-distribution Dataset}}
& \multicolumn{2}{c}{\multirow{2}{*}{\textbf{Average}}} \\
\cmidrule(lr){2-7}
\cmidrule(lr){8-13}

& \multicolumn{2}{c}{\textbf{AVeriTeC}}
& \multicolumn{2}{c}{\textbf{SciFact}}
& \multicolumn{2}{c}{\textbf{QuanTemp++}}
& \multicolumn{2}{c}{\textbf{ClaimPlus}}
& \multicolumn{2}{c}{\textbf{Climate-Fever}}
& \multicolumn{2}{c}{\textbf{HealthFC}}
& \multicolumn{2}{c}{} \\
\cmidrule(lr){2-3}
\cmidrule(lr){4-5}
\cmidrule(lr){6-7}
\cmidrule(lr){8-9}
\cmidrule(lr){10-11}
\cmidrule(lr){12-13}
\cmidrule(l){14-15}
& \textbf{Acc} & \textbf{F1}
& \textbf{Acc} & \textbf{F1}
& \textbf{Acc} & \textbf{F1}
& \textbf{Acc} & \textbf{F1}
& \textbf{Acc} & \textbf{F1}
& \textbf{Acc} & \textbf{F1}
& \textbf{Acc} & \textbf{F1} \\
\midrule

\rowcolor{gray!12}
\multicolumn{15}{@{}c@{}}{
  \itshape\bfseries End-to-end Generation Models
} \\

DeepSeek-V4-Flash
& 58.60 & 40.04
& 76.06 & 67.72
& 49.35 & 40.21
& 49.38 & 38.30
& 45.99 & 32.40
& 44.53 & 46.70
& 53.99 & 44.23 \\

GPT-5.4
& 53.40 & 36.48
& \underline{79.26} & \underline{76.21}
& 44.07 & 37.07
& 51.25 & 41.05
& \textbf{48.60} & 34.30
& \textbf{61.47} & {55.90}
& 56.34 & 46.84 \\

Claude-Sonnet-4.6
& 61.40 & 42.52
& 69.15 & 59.20
& 49.03 & 43.17
& 45.63 & 39.50
& 46.19 & 31.70
& 29.47 & 30.20
& 50.15 & 41.05 \\

Qwen3-8B
& 48.00 & 35.49
& 52.66 & 48.56
& 35.51 & 32.91
& 33.75 & 30.24
& 43.91 & \underline{37.55}
& 40.53 & 45.10
& 42.39 & 38.31 \\

GPT-5.4 w/ Tools
& \textbf{77.00} & \underline{48.50}
& \underline{79.26} & 76.15
& 65.21 & 48.31
& \textbf{66.88} & \textbf{66.94}
& 45.60 & 30.30
& 55.87 & 55.80
& \underline{64.97} & \underline{54.33} \\

\rowcolor{gray!12}
\multicolumn{15}{@{}c@{}}{
  \itshape\bfseries OpenSearch Agents
} \\

MiroThink w/ Tools
& 72.60 & 45.21
& 70.74 & 67.86
& \underline{70.62} & \underline{62.63}
& 58.13 & 50.60
& \underline{47.95} & 35.43
& 42.80 & 45.94
& 60.47 & 51.28 \\

Qwen3-8B w/ Tools
& 57.80 & 42.18
& 68.62 & 62.47
& 49.87 & 48.23
& 55.00 & 46.39
& 42.28 & 35.78
& 30.00 & 36.11
& 50.60 & 45.19 \\

Tongyi-DR w/ Tools
& 55.60 & 40.64
& 70.21 & 67.81
& 56.39 & 50.75
& 53.75 & 46.38
& 42.93 & 31.81
& 35.20 & 32.23
& 52.35 & 44.94 \\

\rowcolor{gray!12}
\multicolumn{15}{@{}c@{}}{
  \itshape\bfseries AFC Systems
} \\

ClaimCheck
& 47.40 & 35.30
& 65.43 & 58.89
& 41.28 & 39.90
& 45.00 & 40.83
& 42.28 & 36.07
& {57.07} & \textbf{56.76}
& 49.74 & 44.63 \\

DEFAME
& 60.00 & 43.99
& 66.49 & 53.78
& 48.48 & 42.64
& 50.63 & 45.60
& 39.09 & 33.71
& 55.87 & 55.44
& 53.43 & 45.86 \\

\rowcolor{gray!12}
\multicolumn{15}{@{}c@{}}{
  \itshape\bfseries Ours
} \\

OpenFC (Cold Start)
& 66.20 & 46.49
& 69.15 & 65.67
& 64.49 & 57.97
& 62.50 & 53.82
& 43.65 & 33.19
& 40.40 & 45.12
& 57.73 & 50.38 \\

OpenFC
& \underline{75.40} & \textbf{52.20}
& \textbf{89.89} & \textbf{84.79}
& \textbf{86.93} & \textbf{84.93}
& \underline{65.63} & \underline{62.17}
& 47.17 & \textbf{39.23}
& \underline{57.33} & \underline{56.50}
& \textbf{70.39} & \textbf{63.30} \\

\bottomrule
\end{tabular*}
}
\caption{
Main results on six fact-checking datasets, including three ID and three OOD benchmarks. We report accuracy and dataset-specific macro-F1. Bold and underline indicate the best and second-best results, respectively.
}
\label{tab:main-results}
\vspace{-6pt}
\end{table*}

\subsection{Experimental Setup}

\paragraph{Datasets.}
The suite includes AVeriTeC~\citep{schlichtkrull2023averitec} for real-world web claims, SciFact~\citep{wadden2020scifact} for scientific article claims, QuanTemp++~\citep{venktesh2025quantempplus} for numerical and temporal reasoning, ClaimPlus~\citep{braun2025defame} for recent broad-domain claims, Climate-FEVER~\citep{diggelmann2020climatefever} for climate claims and HealthFC~\citep{vladika2024healthfc} for health claims.


\paragraph{Metrics.}
We report accuracy and dataset-specific macro-F1 as percentages.
Macro-F1 averages over the label set evaluated for each benchmark: We report macro-F1 over \textsc{Supports}/\textsc{Refutes} for SciFact, as its \textsc{NEI} label set is not directly compatible with our evaluation setting.
We introduce three metrics to analyze the information quality for tools: exact repetition (E-Rep.), semantic repetition at a similarity threshold of 0.95 (S-Rep@0.95), and information gain (Info Gain). E-Rep. measures the proportion of exactly repeated search queries or visited webpages, while S-Rep@0.95 captures semantic redundancy among retrieved contents with similarity above 0.95. Info Gain quantifies the amount of novel information introduced by each retrieval relative to the entire previous corpus.

\paragraph{Comparison Methods.}
We organize the comparison methods into four groups.
\textit{End-to-end generation models} include GPT-5.4, Claude-Sonnet-4.6, and DeepSeek-V4-Flash, evaluated either through direct generation without tools or with the providers' native web-search capabilities.
\textit{OpenSearch agents} include Qwen3-8B, Tongyi-DeepResearch-30B-A3B, MiroThink-30B-A3B, and \textit{OpenFC}. These methods use the same fact-checking prompt, shared \textsc{Search}/\textsc{Visit} environment, and 30-call budget, and therefore provide the controlled policy-level comparison.
\textit{AFC systems} include ClaimCheck and DEFAME, both instantiated with GPT-5.1 as the backend model. 

\paragraph{Training Details.}
The cold-start corpus contains 4,361 calibrated claim-level trajectories and the RL corpus contains 7,483 claims after difficulty filtering, from the training splits of AVeriTeC, QuanTemp++, and SciFact with some overlap between them.
Cold-start SFT fine-tunes Qwen3-8B with full parameters, and RL initializes from that SFT checkpoint and rolls out eight trajectories per sample. The 300-step RL run uses eight NVIDIA A100 GPUs for 4.5 days wall-clock.


\subsection{Main Results}
Table~\ref{tab:main-results} reports verdict quality across three ID and three OOD benchmarks. OpenFC attains the highest overall averages, with 70.39 accuracy and 63.30 macro-F1. Its strongest result is on QuanTemp++, where it reaches 86.93 accuracy and 84.93 F1, exceeding Tongyi-DeepResearch under the same interface by 30.54 and 34.18 points. It also performs strongly on SciFact.
The shared-interface agents provide the controlled policy comparison. Provider-native search, direct generation, and AFC systems differ in their backends or interaction pipelines and are included as heterogeneous end-to-end references rather than policy-only controls. OOD datasets remain more challenging, particularly Climate-Fever and HealthFC. OpenFC nevertheless improves over its Cold Start checkpoint on every reported dataset and metric, raising the overall average by 12.66 accuracy and 12.93 macro-F1 points.
As shown in Table~\ref{tab:tool-usage}, OpenFC uses 9.07 tool calls per sample, a moderate level among the evaluated agents. Compared with OpenFC SFT, it reduces both \textsc{Search} and \textsc{Visit} calls while substantially improving accuracy, suggesting more effective interaction-budget allocation rather than simply more or fewer tool calls.

\begin{table}[t]
\centering
{\footnotesize
\renewcommand{\arraystretch}{1.08}
\setlength{\tabcolsep}{2.0pt}
\begin{tabular*}{\columnwidth}{
  @{\extracolsep{\fill}}lccccc@{}
}
\toprule
\textbf{Method}
& \textbf{Avg Acc.}
& \textbf{Tools}
& \textbf{Visit}
& \textbf{Search}
& \textbf{Auto} \\
\midrule

Qwen3-8B + Tools
& 50.60
& 2.28
& 1.10
& 1.18
& -- \\

ClaimCheck
& 49.74
& 11.72
& --
& 3.08
& 8.65 \\

DEFAME
& 53.43
& 14.91
& --
& 4.27
& 10.64 \\

Tongyi-DR-30B
& 52.35
& 8.90
& 4.24
& 4.67
& -- \\

MiroThink-30B
& 60.47
& 17.47
& 5.03
& 12.44
& -- \\

OpenFC (Cold Start)
& 57.73
& 9.92
& 2.72
& 7.20
& -- \\

OpenFC
& 70.39
& 9.07
& 2.35
& 6.72
& -- \\

\bottomrule
\end{tabular*}
}
\caption{Average tool use per sample.
{Tools} is the total number of explicit \textsc{Search} and \textsc{Visit} calls; {Auto} reports automatic page fetching for ClaimCheck and DEFAME.
}
\vspace{-6pt}
\label{tab:tool-usage}
\end{table}

\begin{table*}[t]
\centering
{\footnotesize
\renewcommand{\arraystretch}{1.10}
\setlength{\tabcolsep}{2pt}
\begin{tabular*}{\textwidth}{
  @{\extracolsep{\fill}}l*{11}{c}@{}
}
\toprule
\multirow{2}{*}{\textbf{Method}}
& \multicolumn{2}{c}{\textbf{AVeriTeC}}
& \multicolumn{2}{c}{\textbf{SciFact}}
& \multicolumn{2}{c}{\textbf{QuanTemp++}}
& \multicolumn{2}{c}{\textbf{ClaimPlus}}
& \multicolumn{3}{c}{\textbf{Tool Usage}} \\
\cmidrule(lr){2-3}
\cmidrule(lr){4-5}
\cmidrule(lr){6-7}
\cmidrule(lr){8-9}
\cmidrule(lr){10-12}
& \textbf{Acc.} & \textbf{F1}
& \textbf{Acc.} & \textbf{F1}
& \textbf{Acc.} & \textbf{F1}
& \textbf{Acc.} & \textbf{F1}
& \textbf{Tools}
& \textbf{Search}
& \textbf{Visit} \\
\midrule

Vanilla SFT
& 65.80 & 43.14
& 63.30 & 62.55
& 61.51 & 55.53
& 60.00 & 48.95
& 9.89 & 5.08 & 4.81 \\

Stepwise-Calibration SFT
& 66.20 & \underline{46.49}
& 69.15 & 65.67
& 64.49 & 57.97
& \underline{62.50} & \underline{53.82}
& 9.92 & 7.20 & 2.72 \\

\midrule

Vanilla GRPO
& 71.60 & 37.80
& 81.38 & 71.75
& 73.61 & 68.90
& 52.50 & 35.84
& 8.02 & 5.99 & 2.02 \\

\quad + Advantage reweighting
& \underline{74.00} & 41.37
& \underline{88.83} & \underline{83.31}
& \underline{86.28} & \underline{82.91}
& 58.75 & 40.88
& 6.43 & 4.33 & 2.11 \\

\quad + Tool reward
& 70.60 & 40.68
& \underline{88.83} & 82.03
& 85.73 & 82.57
& 60.00 & 45.13
& 8.20 & 7.61 & 0.59 \\

Verification-aware RL
& \textbf{75.40} & \textbf{52.20}
& \textbf{89.89} & \textbf{84.79}
& \textbf{86.93} & \textbf{84.93}
& \textbf{65.63} & \textbf{62.17}
& 9.07 & 6.72 & 2.35 \\

\bottomrule
\end{tabular*}
}
\caption{
Ablation study of the proposed training components.
The best and second-best task-performance results are highlighted in bold
and underlined, respectively.
}
\label{tab:openfc_ablation}
\vspace{-6pt}
\end{table*}

\subsection{Ablation Studies}
Table~\ref{tab:openfc_ablation} isolates the effects of stepwise calibration and verification-aware RL using single-run configurations.

\paragraph{Ablation on Stepwise Calibration.}
Vanilla SFT follows the same cold-start data setting but directly trains on trajectories rolled out by Tongyi-DR, without stepwise supervisor intervention. Stepwise-Calibration SFT improves all eight verdict metrics, with particularly clear gains on SciFact and ClaimPlus, while leaving the total tool budget nearly unchanged (9.92 vs.\ 9.89 calls). It also shifts usage from \textsc{Visit} to \textsc{Search}, suggesting that calibration improves evidence acquisition before proposed actions affect the subsequent trajectory. The supervisor remains label-free, as correctness filtering is applied only after rollout termination.

\paragraph{Ablation on Verification-Aware RL.}
Vanilla GRPO improves accuracy on all three ID benchmarks but yields uneven macro-F1 and substantially degrades ClaimPlus relative to the calibrated SFT checkpoint. Advantage reweighting strengthens performance, especially on SciFact and QuanTemp++, whereas the tool reward produces a more search-oriented interaction pattern and improves ClaimPlus. Their combination achieves the best result on all four benchmarks: compared with Vanilla GRPO, Verification-aware RL improves accuracy by 3.80--13.32 points and macro-F1 by 13.04--26.33 points, while maintaining a moderate budget of 9.07 tool calls. The format reward is evaluated for every rollout. The bounded tool reward is computed only when the rollout group contains an eligible correct rollout that defines $t_{\min}$, and it is skipped for all-wrong groups.

\subsection{Further Analysis}

\begin{figure}[t]
\centering
\includegraphics[width=1\columnwidth]{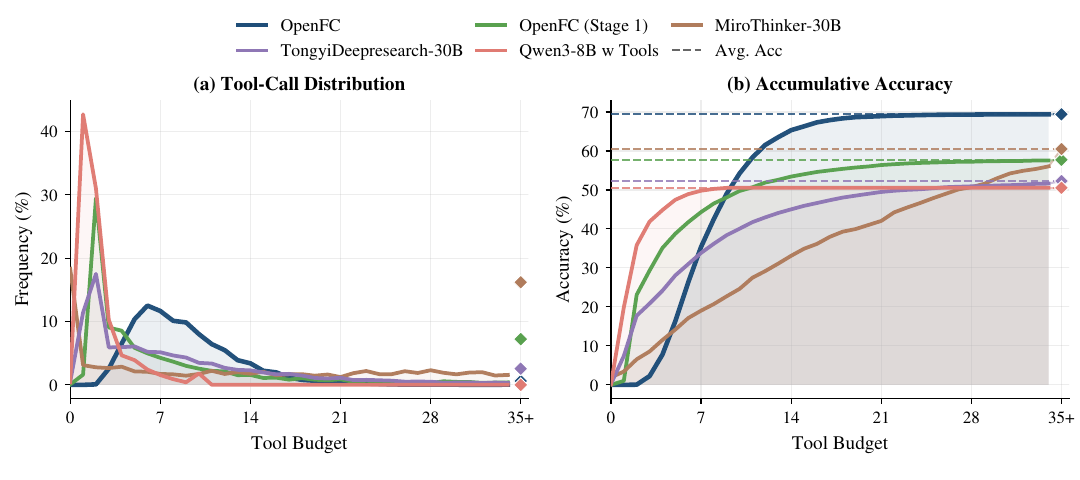}
\caption{Tool-call distributions and cumulative accuracy under different call budgets.}
\label{fig:tool-call-diagnostics}
\end{figure}

\begin{table}[t]
\centering
{\footnotesize
\renewcommand{\arraystretch}{1.08}
\setlength{\tabcolsep}{2.0pt}
\begin{tabular*}{\columnwidth}{
  @{\extracolsep{\fill}}lccc@{}
}
\toprule
\textbf{Method}
& \textbf{E-Rep.} $\downarrow$
& \textbf{S-Rep@0.95} $\downarrow$
& \textbf{Info Gain} $\uparrow$ \\
\midrule
OpenFC 
& \textbf{0.19\%}
& 3.33\%
& 14.67\% \\
OpenFC (Cold Start)
& 8.27\%
& 11.24\%
& 13.68\% \\

Qwen3-8B w/ Tools
& 1.08\%
& \textbf{1.42\%}
& 12.79\% \\

Tongyi-DR
& 0.72\%
& 4.19\%
& \textbf{15.58\%} \\

MiroThink
& 1.12\%
& 9.40\%
& 10.73\% \\
\bottomrule
\end{tabular*}
}
\caption{Analysis on information quality across different methods on QuanTemp++.}
\label{tab:search_behavior}
\vspace{-6pt}
\end{table}

\paragraph{Information Quality Analysis.}
Relative to OpenFC SFT, the final policy reduces exact repetition from 8.27\% to 0.19\% and semantic repetition from 11.24\% to 3.33\%, while increasing information gain from 13.68\% to 14.67\% (Table~\ref{tab:search_behavior}). RL therefore changes query reformulation as well as search frequency. OpenFC combines high novelty with the lowest exact repetition, matching a verifier that uses new queries to resolve remaining evidentiary uncertainty rather than pursue diversity alone.

\paragraph{Tool Distributions}
Figure~\ref{fig:tool-call-diagnostics} complements average counts with the full budget distribution. Qwen3-8B and OpenFC (Cold Start) concentrate more trajectories at small budgets, whereas OpenFC shifts mass toward moderate-depth verification without the pronounced long tail of MiroThinker. Together with the cumulative-accuracy curves, this characterizes a policy that continues verification when additional interactions are useful while keeping its mean budget below the SFT checkpoint.

\begin{figure}[t]
\centering
\includegraphics[width=\columnwidth]{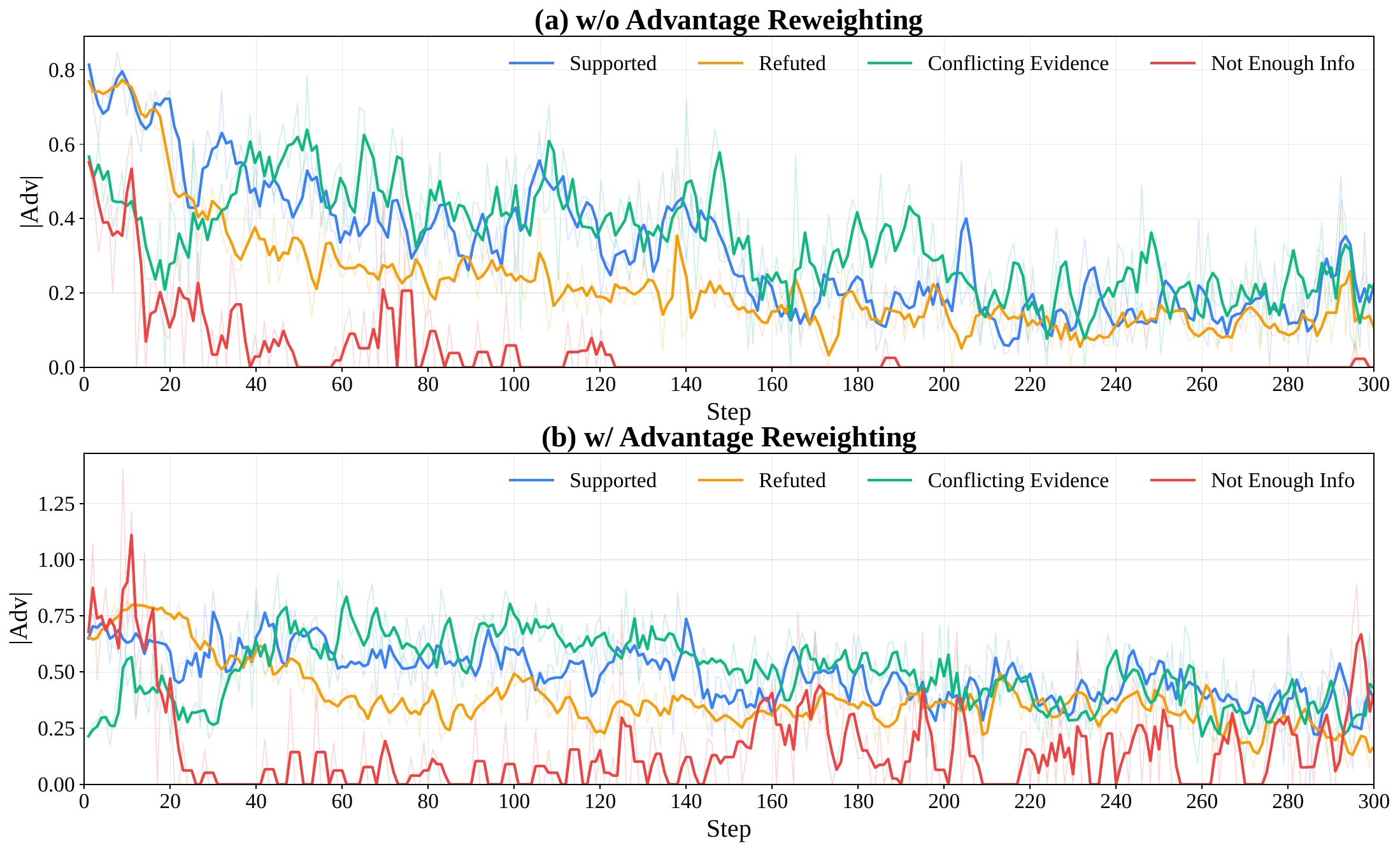}
\caption{Label-level absolute advantages with and without reweighting.}
\label{fig:advantage-reweighting}
\vspace{-6pt}
\end{figure}

\paragraph{Training Dynamics}

Figure~\ref{fig:advantage-reweighting} shows that the \textsc{NEI} advantage approaches zero without reweighting but remains active with it. Reweighting thus preserves learning pressure for a verdict that requires stopping without forcing a supported or refuted decision when evidence remains insufficient, consistent with the complete configuration's stronger macro-F1.

\section{Conclusion}
OpenFC combines Stepwise-Calibrated Cold Start with Verification-Aware Reinforcement Learning to learn reasoning, evidence acquisition, and stopping under a unified interface. SCCS provides label-free pre-execution calibration, while VA-RL couples universal format feedback with conditional tool rewards and localized masking that separates external execution failures from policy errors. Across six benchmarks, OpenFC demonstrates a clear overall advantage over existing methods, while also consistently improving upon its cold-start policy. Future work should further strengthen robustness in these challenging settings and extend the verification policy to stronger backbones and richer evidence sources while preserving controlled tool use.


\bibliography{aaai2027}

\ifnum\includeOpenFCAppendix=1
\appendix
\setcounter{secnumdepth}{1}
\twocolumn[
\begin{center}
    {\LARGE\bfseries OpenFC: Learning Verification Policies for Open-Search Fact Checking\par}
    \vspace{0.75em}
    {\large\bfseries Technical Supplement\par}
    \vspace{1em}
\end{center}
]
\suppressfloats[t]
\section{Dataset and Metrics}
\label{app:data-metric}

\subsection{Dataset Details}
We divide the benchmarks by their role in our experiments.
In-distribution datasets provide training claims and disjoint evaluation claims, whereas out-of-distribution datasets are used only for evaluation.

\subsubsection{In-Distribution Datasets}

\paragraph{AVeriTeC.}
AVeriTeC contains 4,568 real-world claims collected from 50 professional fact-checking organizations and pairs each claim with web-sourced question--answer evidence and a textual justification~\citep{schlichtkrull2023averitec}.
Its temporal split contains 3,068 training, 500 development, and 1,000 test claims.
The four canonical verdicts are \textsc{Supported}, \textsc{Refuted}, \textsc{Conflicting Evidence/Cherrypicking}, and \textsc{NEI}, where \textsc{NEI} denotes \textsc{Not Enough Information}.
The training split is imbalanced toward \textsc{Refuted}: it contains 847 supported, 1,743 refuted, 196 conflicting, and 282 NEI claims.
The preprocessing manifest used by all training stages contains 3,067 of the official 3,068 training claims. The single record absent from this manifest is excluded consistently from trajectory construction and difficulty filtering, so every reported AVeriTeC training statistic uses 3,067 as its denominator; the complete 500-claim development evaluation split is unaffected.

\paragraph{QuanTemp++.}
QuanTemp++ targets naturally occurring numerical claims whose verification may require temporal, interval, statistical, or comparison reasoning~\citep{venktesh2025quantempplus}.
It contains 9,935 training, 3,084 validation, and 2,495 test claims, together with an open-domain collection of approximately 165.7K evidence records.
Its native verdicts are \textsc{True}, \textsc{False}, and \textsc{Conflicting}, which correspond to OpenFC's \textsc{Supported}, \textsc{Refuted}, and \textsc{Conflicting Evidence} outputs.
The benchmark filters fact-checking pages and evidence published after the claim date, making it particularly relevant to open-search verification without gold-answer or temporal leakage.

\paragraph{SciFact.}
SciFact consists of 1,409 expert-written scientific claims verified against 5,183 research abstracts~\citep{wadden2020scifact}.
The official train/dev/test split is 809/300/300; the corresponding class totals are 556 \textsc{Supports}, 337 \textsc{Refutes}, and 516 \textsc{NoInfo}.
Importantly, \textsc{NoInfo} means that the associated abstract corpus contains no annotated support or refutation relation.
It is therefore corpus-relative and does not establish that evidence is unavailable on the open web.
We directly transfer only the two evidence-grounded labels, \textsc{Supports} and \textsc{Refutes}, as SciFact gold labels.
All SciFact \textsc{NoInfo} instances are excluded from both training and evaluation; we do not relabel them using open-web evidence.

\subsubsection{Out-of-Distribution Datasets}

\paragraph{ClaimPlus.}
ClaimPlus is our text-only evaluation subset of CLAIMREVIEW2024+, introduced with DEFAME~\citep{braun2025defame}.
The source benchmark contains 300 English claims retrieved through the Google FactCheck Claim Search API, comprising 160 text-only and 140 text--image claims dated between November 2023 and January 2025.
Across all 300 source claims, the label distribution is 89 \textsc{Supported}, 129 \textsc{Refuted}, 61 \textsc{Misleading}, and 21 \textsc{NEI}.
We evaluate all 160 text-only claims; the 140 text--image claims are excluded. The aggregate counts for the 300-claim collection do not uniquely determine the per-label distribution of this text-only subset, so we do not infer subset-level counts without the evaluation manifest.
These claims are not exposed during training, thereby testing recent, broad-domain generalization under the same text-and-web-search interface used by the other benchmarks.

\paragraph{Climate-FEVER.}
Climate-FEVER contains 1,535 real-world climate claims and five Wikipedia evidence candidates per claim, for 7,675 claim--evidence pairs~\citep{diggelmann2020climatefever}.
The aggregate claim labels are 655 \textsc{Supports}, 253 \textsc{Refutes}, 153 \textsc{Disputed}, and 474 \textsc{NEI}.
Unlike FEVER-style three-way datasets, \textsc{Disputed} explicitly captures claims for which both supporting and refuting evidence was found.
We use Climate-FEVER only as an OOD climate-domain evaluation set.

\paragraph{HealthFC.}
HealthFC contains 750 health-related claims in parallel German and English versions, with medical-expert verdicts and evidence drawn from systematic reviews and clinical trials~\citep{vladika2024healthfc}.
The dataset includes 202 \textsc{Supported}, 125 \textsc{Refuted}, and 423 \textsc{NEI} claims; supported and refuted instances additionally carry low, medium, or high evidence-strength annotations.

\subsection{Evaluation Metrics}

\subsubsection{Verdict metrics.}
We report verdict accuracy and macro-F1 as percentages, computed over the
evaluation label set $\mathcal{Y}$ of each dataset after the canonicalization
described below.

\paragraph{Accuracy.}
Accuracy is the number of correctly classified claims divided by the total
number of evaluated claims. It is dominated by the majority verdicts of a
dataset, which is why we always report it together with macro-F1.

\paragraph{Macro-F1.}
For each class $y\in\mathcal{Y}$ we first compute
\begin{equation}
\begin{aligned}
P_y  &=\frac{\mathrm{TP}_y}{\mathrm{TP}_y+\mathrm{FP}_y},\\
R_y  &=\frac{\mathrm{TP}_y}{\mathrm{TP}_y+\mathrm{FN}_y},\\
F1_y &=\frac{2P_yR_y}{P_y+R_y},
\end{aligned}
\end{equation}
and then give every verdict class equal weight, irrespective of its frequency:
\begin{equation}
\operatorname{MacroF1}
=\frac{1}{|\mathcal Y|}
\sum_{y\in\mathcal Y}F1_y.
\end{equation}
We set a class-level precision, recall, or F1 term to zero when its
denominator is zero.

\paragraph{Label canonicalization.}
Dataset-specific names are canonicalized before scoring:
\textsc{True}/\textsc{False} map to \textsc{Supported}/\textsc{Refuted}, and
\textsc{Conflicting}, \textsc{Disputed}, and \textsc{Misleading} map to the
benchmark adapter's \textsc{Conflicting Evidence} output.
\textsc{NoInfo} and dataset-specific not-enough-information labels map to
\textsc{NEI} for datasets that contain this class.
SciFact is evaluated only over \textsc{Supports} and \textsc{Refutes}; its
\textsc{NoInfo} instances are excluded.
The resulting label set is therefore two classes for SciFact, three for
QuanTemp++ and HealthFC, and each remaining benchmark's native verdict set.

\subsubsection{Information quality metrics.}
Verdict scores say nothing about how the evidence was obtained, so we
additionally measure the quality of the retrieval process and report all three
metrics as percentages averaged over trajectories.
We linearize a trajectory into $T$ search rounds ordered by time, where one
round is one search tool call: round $t$ issues the queries in $Q_t$ and
returns a set of evidence items, each item being the title and the snippet of
one retrieved result.
Evidence strings shorter than $50$ characters are discarded as uninformative,
and only evidence returned by search calls is used, since the format and the
success behavior of explicitly visited pages differ across systems.
We write $\mathrm{sim}(\cdot,\cdot)$ for the cosine similarity between the
\texttt{bge-m3} embeddings of two texts.

\paragraph{Exact query repetition (E-Rep.).}
E-Rep.\ is the fraction of queries that repeat an earlier query verbatim.
Let $q_1,\dots,q_N$ be the queries of a trajectory in temporal order, each
normalized by collapsing whitespace and case-folding, with empty strings
removed:
\begin{equation}
\operatorname{E\text{-}Rep}
=1-\frac{\left|\{q_1,\dots,q_N\}\right|}{N}.
\end{equation}
Duplication is assessed against the whole trajectory rather than against the
immediately preceding call only, so a query that reappears many rounds later
is still counted. Trajectories without any usable query are excluded.

\paragraph{Semantic query repetition (S-Rep@$0.95$).}
E-Rep.\ only detects literal duplicates, while an agent may equally well
re-ask the same question in different words. We therefore count a query as a
semantic repetition when it is close to \emph{any} query issued in a strictly
earlier round:
\begin{equation}
\operatorname{S\text{-}Rep}@0.95
=\frac{1}{N}\sum_{t=1}^{T}\sum_{q\in Q_t}
\mathbf{1}\!\left[\max_{s<t,\;q'\in Q_s}\mathrm{sim}(q,q')\ \ge\ 0.95\right].
\end{equation}
The threshold is deliberately high, so a counted query is a near-paraphrase of
one already issued rather than a legitimate refinement of it. Queries issued
in the same round do not serve as history for one another, and the maximum
over an empty history is $-\infty$, so the first round enters the denominator
without ever contributing to the numerator.

\paragraph{Information gain (Info Gain).}
Avoiding repeated queries is not sufficient, since distinct queries can still
return overlapping content, so Info Gain measures redundancy on the evidence
side. Let $E_1,\dots,E_K$ be the non-empty evidence sets of a trajectory in
temporal order. Every item of round $k$ is scored against all evidence
retained so far, and the round scores are then combined:
\begin{equation}
\begin{aligned}
g_k &=\frac{1}{|E_k|}\sum_{e\in E_k}
\Bigl(1-\max_{e'\in E_1\cup\dots\cup E_{k-1}}\mathrm{sim}(e,e')\Bigr),\\
\operatorname{InfoGain} &=\frac{1}{K-1}\sum_{k=2}^{K}g_k .
\end{aligned}
\end{equation}
Averaging within a round before averaging across rounds keeps a single
result-rich call from dominating the trajectory.

\section{More Results}
\label{app:more-results}

To complement the aggregate accuracy and macro-F1 results in the main paper, Tables~\ref{tab:per-label-averitec}--\ref{tab:per-label-healthfc} report precision, recall, and F1 for each evaluated verdict label. The Overall columns repeat the dataset-level accuracy and macro-F1 from the main table. The evaluated label set follows each benchmark's native verdict space after canonicalization: SciFact uses two labels; QuanTemp++ and HealthFC use three labels; and AVeriTeC, ClaimPlus, and Climate-Fever use four labels. We use SUP for supported, REF for refuted, NEI for not enough information, and CON for conflicting evidence.

The label-level results show that OpenFC's improvements are not limited to the dominant binary verdicts, but extend to the more challenging insufficient- and conflicting-evidence categories. On the in-distribution datasets, reinforcement learning improves both labels on SciFact, raising the \textsc{Supported} and \textsc{Refuted} F1 scores from 72.23/59.11 to 88.01/81.57, respectively. The largest change appears on QuanTemp++, where the \textsc{Conflicting Evidence} F1 increases from 33.23 to 75.88, alongside substantial gains for \textsc{Supported} and \textsc{Refuted}. On AVeriTeC, OpenFC also improves all four label-level F1 scores over the cold-start policy, producing a more balanced verifier rather than relying primarily on the frequent support and refute decisions. This contrasts with several strong baselines that achieve high precision but extremely low recall on \textsc{NEI} or \textsc{Conflicting Evidence}, indicating that these labels are rarely predicted. The gains transfer partially to out-of-distribution data: OpenFC substantially improves \textsc{NEI} on ClaimPlus, all three labels on HealthFC, and the \textsc{Supported}/\textsc{Refuted} classes on Climate-FEVER. Nevertheless, \textsc{NEI} performance decreases on Climate-FEVER and \textsc{Conflicting Evidence} remains difficult on ClaimPlus, suggesting that abstention and conflict recognition are still sensitive to domain shift. Overall, the results indicate that verification-aware RL improves class balance and fine-grained verdict discrimination, while leaving robust recognition of ambiguous evidence as an important remaining challenge.

\begin{table*}[t]
\centering
{\footnotesize
\renewcommand{\arraystretch}{1.10}
\setlength{\tabcolsep}{2pt}
\begin{tabular*}{\textwidth}{@{\extracolsep{\fill}}l*{14}{c}@{}}
\toprule
\multicolumn{1}{c}{\multirow{2}{*}{\textbf{Method}}}
& \multicolumn{3}{c}{\textbf{SUP}}
& \multicolumn{3}{c}{\textbf{REF}}
& \multicolumn{3}{c}{\textbf{NEI}}
& \multicolumn{3}{c}{\textbf{CON}}
& \multicolumn{2}{c}{\textbf{Overall}} \\
\cmidrule(lr){2-4}
\cmidrule(lr){5-7}
\cmidrule(lr){8-10}
\cmidrule(lr){11-13}
\cmidrule(l){14-15}
& \textbf{P} & \textbf{R} & \textbf{F1} & \textbf{P} & \textbf{R} & \textbf{F1} & \textbf{P} & \textbf{R} & \textbf{F1} & \textbf{P} & \textbf{R} & \textbf{F1} & \textbf{Acc} & \textbf{F1} \\
\midrule

\rowcolor{gray!12}
\multicolumn{15}{@{}c@{}}{
  \itshape\bfseries End-to-end Generation Models
} \\

DeepSeek-V4-Flash
& 47.87 & 81.15 & 60.22 & 80.54 & 60.65 & 69.19 & 16.49 & 14.29 & 15.31 & 28.96 & 10.53 & 15.44
& 58.60 & 40.04 \\

GPT-5.4
& 49.20 & 85.20 & 62.38 & 86.27 & 50.45 & 63.67 & 10.36 & 23.08 & 14.30 & 96.03 & 2.87 & 5.57
& 53.40 & 36.48 \\

Claude-Sonnet-4.6
& 56.37 & 65.68 & 60.67 & 82.73 & 69.46 & 75.52 & 13.91 & 25.74 & 18.06 & 15.85 & 15.82 & 15.83
& 61.40 & 42.52 \\

Qwen3-8B
& 58.14 & 35.36 & 43.97 & 70.10 & 57.98 & 63.47 & 9.43 & 37.17 & 15.04 & 20.64 & 18.45 & 19.48
& 48.00 & 35.49 \\

GPT-5.4 w/ Tools
& 74.73 & 82.45 & 78.40 & 78.14 & 90.48 & 83.86 & 47.23 & 20.34 & 28.43 & 3.17 & 3.47 & 3.31
& 77.00 & 48.50 \\

\rowcolor{gray!12}
\multicolumn{15}{@{}c@{}}{
  \itshape\bfseries OpenSearch Agents
} \\

MiroThink w/ Tools
& 67.76 & 58.12 & 62.57 & 83.46 & 93.24 & 88.08 & 15.39 & 10.36 & 12.38 & 51.50 & 10.77 & 17.81
& 72.60 & 45.21 \\

Qwen3-8B w/ Tools
& 67.63 & 67.16 & 67.39 & 83.88 & 62.59 & 71.69 & 11.13 & 23.08 & 15.02 & 11.13 & 21.29 & 14.62
& 57.80 & 42.18 \\

Tongyi-DR w/ Tools
& 59.65 & 76.34 & 66.97 & 87.85 & 58.31 & 70.09 & 21.51 & 11.46 & 14.95 & 15.81 & 7.92 & 10.55
& 55.60 & 40.64 \\

\rowcolor{gray!12}
\multicolumn{15}{@{}c@{}}{
  \itshape\bfseries AFC Systems
} \\

ClaimCheck
& 72.40 & 41.09 & 52.43 & 81.47 & 54.05 & 64.99 & 9.34 & 57.17 & 16.06 & 14.31 & 5.29 & 7.72
& 47.40 & 35.30 \\

DEFAME
& 79.56 & 46.69 & 58.85 & 80.03 & 71.97 & 75.79 & 15.41 & 54.99 & 24.07 & 36.58 & 11.29 & 17.25
& 60.00 & 43.99 \\

\rowcolor{gray!12}
\multicolumn{15}{@{}c@{}}{
  \itshape\bfseries Ours
} \\

OpenFC (Cold Start)
& 67.53 & 76.43 & 71.70 & 82.69 & 73.38 & 77.76 & 9.37 & 12.66 & 10.77 & 26.49 & 25.02 & 25.73
& 66.20 & 46.49 \\

OpenFC
& 84.18 & 74.59 & 79.10 & 79.89 & 89.18 & 84.28 & 57.42 & 11.43 & 19.06 & 26.40 & 26.32 & 26.36
& 75.40 & 52.20 \\

\bottomrule
\end{tabular*}
}
\caption{Label-level results on AVeriTeC. The labels in the dataset are Supported / Refuted / NEI / Conflicting Evidence.}
\label{tab:per-label-averitec}
\end{table*}

\begin{table*}[t]
\centering
{\footnotesize
\renewcommand{\arraystretch}{1.10}
\setlength{\tabcolsep}{2pt}
\begin{tabular*}{\textwidth}{@{\extracolsep{\fill}}l*{8}{c}@{}}
\toprule
\multicolumn{1}{c}{\multirow{2}{*}{\textbf{Method}}}
& \multicolumn{3}{c}{\textbf{SUP}}
& \multicolumn{3}{c}{\textbf{REF}}
& \multicolumn{2}{c}{\textbf{Overall}} \\
\cmidrule(lr){2-4}
\cmidrule(lr){5-7}
\cmidrule(l){8-9}
& \textbf{P} & \textbf{R} & \textbf{F1} & \textbf{P} & \textbf{R} & \textbf{F1} & \textbf{Acc} & \textbf{F1} \\
\midrule

\rowcolor{gray!12}
\multicolumn{9}{@{}c@{}}{
  \itshape\bfseries End-to-end Generation Models
} \\

DeepSeek-V4-Flash
& 53.44 & 80.90 & 64.36 & 70.94 & 71.22 & 71.08
& 76.06 & 67.72 \\

GPT-5.4
& 71.62 & 84.10 & 77.36 & 75.71 & 74.42 & 75.06
& 79.26 & 76.21 \\

Claude-Sonnet-4.6
& 49.77 & 64.32 & 56.12 & 53.78 & 73.98 & 62.28
& 69.15 & 59.20 \\

Qwen3-8B
& 62.56 & 51.57 & 56.54 & 32.23 & 54.76 & 40.58
& 52.66 & 48.56 \\

GPT-5.4 w/ Tools
& 67.99 & 84.10 & 75.19 & 80.00 & 74.42 & 77.11
& 79.26 & 76.15 \\

\rowcolor{gray!12}
\multicolumn{9}{@{}c@{}}{
  \itshape\bfseries OpenSearch Agents
} \\

MiroThink w/ Tools
& 67.82 & 75.98 & 71.67 & 67.94 & 60.59 & 64.05
& 70.74 & 67.86 \\

Qwen3-8B w/ Tools
& 58.07 & 73.56 & 64.90 & 61.06 & 59.05 & 60.04
& 68.62 & 62.47 \\

Tongyi-DR w/ Tools
& 70.66 & 70.16 & 70.41 & 60.80 & 70.31 & 65.21
& 70.21 & 67.81 \\

\rowcolor{gray!12}
\multicolumn{9}{@{}c@{}}{
  \itshape\bfseries AFC Systems
} \\

ClaimCheck
& 56.51 & 66.31 & 61.02 & 51.17 & 63.73 & 56.76
& 65.43 & 58.89 \\

DEFAME
& 58.08 & 60.57 & 59.30 & 34.95 & 77.96 & 48.26
& 66.49 & 53.78 \\

\rowcolor{gray!12}
\multicolumn{9}{@{}c@{}}{
  \itshape\bfseries Ours
} \\

OpenFC (Cold Start)
& 74.42 & 70.16 & 72.23 & 52.77 & 67.19 & 59.11
& 69.15 & 65.67 \\

OpenFC
& 83.74 & 92.74 & 88.01 & 78.94 & 84.38 & 81.57
& 89.89 & 84.79 \\

\bottomrule
\end{tabular*}
}
\caption{Label-level results on SciFact. The labels in the dataset are Supported / Refuted.}
\label{tab:per-label-scifact}
\end{table*}

\begin{table*}[t]
\centering
{\footnotesize
\renewcommand{\arraystretch}{1.10}
\setlength{\tabcolsep}{2pt}
\begin{tabular*}{\textwidth}{@{\extracolsep{\fill}}l*{11}{c}@{}}
\toprule
\multicolumn{1}{c}{\multirow{2}{*}{\textbf{Method}}}
& \multicolumn{3}{c}{\textbf{SUP}}
& \multicolumn{3}{c}{\textbf{REF}}
& \multicolumn{3}{c}{\textbf{CON}}
& \multicolumn{2}{c}{\textbf{Overall}} \\
\cmidrule(lr){2-4}
\cmidrule(lr){5-7}
\cmidrule(lr){8-10}
\cmidrule(l){11-12}
& \textbf{P} & \textbf{R} & \textbf{F1} & \textbf{P} & \textbf{R} & \textbf{F1} & \textbf{P} & \textbf{R} & \textbf{F1} & \textbf{Acc} & \textbf{F1} \\
\midrule

\rowcolor{gray!12}
\multicolumn{12}{@{}c@{}}{
  \itshape\bfseries End-to-end Generation Models
} \\

DeepSeek-V4-Flash
& 33.56 & 78.33 & 46.99 & 78.08 & 56.12 & 65.30 & 39.88 & 4.66 & 8.34
& 49.35 & 40.21 \\

GPT-5.4
& 35.54 & 82.21 & 49.63 & 81.80 & 47.12 & 59.80 & 82.10 & 0.90 & 1.78
& 44.07 & 37.07 \\

Claude-Sonnet-4.6
& 34.99 & 64.46 & 45.36 & 77.76 & 57.96 & 66.42 & 45.74 & 11.00 & 17.73
& 49.03 & 43.17 \\

Qwen3-8B
& 33.73 & 23.06 & 27.39 & 66.04 & 49.44 & 56.55 & 30.88 & 9.72 & 14.79
& 35.51 & 32.91 \\

GPT-5.4 w/ Tools
& 49.50 & 73.57 & 59.18 & 60.62 & 83.27 & 70.16 & 48.62 & 9.28 & 15.59
& 65.21 & 48.31 \\

\rowcolor{gray!12}
\multicolumn{12}{@{}c@{}}{
  \itshape\bfseries OpenSearch Agents
} \\

MiroThink w/ Tools
& 85.66 & 90.32 & 87.93 & 99.99 & 79.32 & 88.46 & 52.52 & 6.46 & 11.50
& 70.62 & 62.63 \\

Qwen3-8B w/ Tools
& 47.16 & 45.15 & 46.13 & 79.65 & 58.62 & 67.54 & 31.21 & 30.84 & 31.02
& 49.87 & 48.23 \\

Tongyi-DR w/ Tools
& 47.51 & 79.84 & 59.57 & 87.92 & 65.55 & 75.10 & 56.88 & 10.40 & 17.58
& 56.39 & 50.75 \\

\rowcolor{gray!12}
\multicolumn{12}{@{}c@{}}{
  \itshape\bfseries AFC Systems
} \\

ClaimCheck
& 52.85 & 32.69 & 40.39 & 83.85 & 57.24 & 68.04 & 40.22 & 6.55 & 11.27
& 41.28 & 39.90 \\

DEFAME
& 48.40 & 32.05 & 38.56 & 74.52 & 68.21 & 71.23 & 54.76 & 10.86 & 18.13
& 48.48 & 42.64 \\

\rowcolor{gray!12}
\multicolumn{12}{@{}c@{}}{
  \itshape\bfseries Ours
} \\

OpenFC (Cold Start)
& 59.55 & 56.93 & 58.21 & 82.59 & 82.35 & 82.47 & 55.35 & 23.74 & 33.23
& 64.49 & 57.97 \\

OpenFC
& 84.10 & 91.57 & 87.68 & 89.96 & 92.53 & 91.23 & 86.30 & 67.71 & 75.88
& 86.93 & 84.93 \\

\bottomrule
\end{tabular*}
}
\caption{Label-level results on QuanTemp++. The labels in the dataset are Supported / Refuted / Conflicting Evidence.}
\label{tab:per-label-quantemp}
\end{table*}

\begin{table*}[t]
\centering
{\footnotesize
\renewcommand{\arraystretch}{1.10}
\setlength{\tabcolsep}{2pt}
\begin{tabular*}{\textwidth}{@{\extracolsep{\fill}}l*{14}{c}@{}}
\toprule
\multicolumn{1}{c}{\multirow{2}{*}{\textbf{Method}}}
& \multicolumn{3}{c}{\textbf{SUP}}
& \multicolumn{3}{c}{\textbf{REF}}
& \multicolumn{3}{c}{\textbf{NEI}}
& \multicolumn{3}{c}{\textbf{CON}}
& \multicolumn{2}{c}{\textbf{Overall}} \\
\cmidrule(lr){2-4}
\cmidrule(lr){5-7}
\cmidrule(lr){8-10}
\cmidrule(lr){11-13}
\cmidrule(l){14-15}
& \textbf{P} & \textbf{R} & \textbf{F1} & \textbf{P} & \textbf{R} & \textbf{F1} & \textbf{P} & \textbf{R} & \textbf{F1} & \textbf{P} & \textbf{R} & \textbf{F1} & \textbf{Acc} & \textbf{F1} \\
\midrule

\rowcolor{gray!12}
\multicolumn{15}{@{}c@{}}{
  \itshape\bfseries End-to-end Generation Models
} \\

DeepSeek-V4-Flash
& 55.35 & 77.09 & 64.44 & 53.86 & 66.68 & 59.59 & 14.41 & 15.38 & 14.88 & 36.46 & 8.89 & 14.29
& 49.38 & 38.30 \\

GPT-5.4
& 62.46 & 86.71 & 72.61 & 57.53 & 60.72 & 59.08 & 18.13 & 46.50 & 26.09 & 49.45 & 3.43 & 6.42
& 51.25 & 41.05 \\

Claude-Sonnet-4.6
& 60.01 & 64.58 & 62.21 & 51.21 & 57.41 & 54.13 & 13.80 & 30.77 & 19.05 & 41.37 & 15.56 & 22.61
& 45.63 & 39.50 \\

Qwen3-8B
& 76.49 & 27.08 & 40.00 & 48.48 & 59.26 & 53.33 & 6.25 & 30.77 & 10.39 & 38.45 & 11.11 & 17.24
& 33.75 & 30.24 \\

GPT-5.4 w/ Tools
& 100.00 & 100.00 & 100.00 & 97.56 & 87.61 & 92.32 & 72.41 & 36.84 & 48.83 & 100.00 & 15.35 & 26.61
& 66.88 & 66.94 \\

\rowcolor{gray!12}
\multicolumn{15}{@{}c@{}}{
  \itshape\bfseries OpenSearch Agents
} \\

MiroThink w/ Tools
& 70.99 & 67.00 & 68.94 & 67.73 & 76.77 & 71.97 & 51.44 & 39.89 & 44.93 & 11.22 & 31.57 & 16.56
& 58.13 & 50.60 \\

Qwen3-8B w/ Tools
& 72.90 & 70.05 & 71.45 & 60.21 & 69.98 & 64.73 & 12.28 & 15.73 & 13.79 & 39.59 & 32.32 & 35.59
& 55.00 & 46.39 \\

Tongyi-DR w/ Tools
& 59.97 & 86.72 & 70.91 & 58.40 & 66.27 & 62.09 & 60.73 & 23.43 & 33.81 & 38.83 & 12.32 & 18.71
& 53.75 & 46.38 \\

\rowcolor{gray!12}
\multicolumn{15}{@{}c@{}}{
  \itshape\bfseries AFC Systems
} \\

ClaimCheck
& 74.05 & 41.67 & 53.33 & 64.27 & 66.67 & 65.45 & 17.90 & 92.31 & 29.99 & 40.06 & 8.89 & 14.55
& 45.00 & 40.83 \\

DEFAME
& 86.72 & 56.12 & 68.14 & 50.22 & 70.71 & 58.73 & 24.40 & 85.15 & 37.93 & 49.27 & 10.71 & 17.60
& 50.63 & 45.60 \\

\rowcolor{gray!12}
\multicolumn{15}{@{}c@{}}{
  \itshape\bfseries Ours
} \\

OpenFC (Cold Start)
& 79.51 & 83.70 & 81.55 & 53.26 & 82.88 & 64.85 & 59.66 & 24.31 & 34.54 & 48.85 & 26.47 & 34.34
& 62.50 & 53.82 \\

OpenFC
& 90.76 & 77.08 & 83.36 & 53.72 & 98.16 & 69.44 & 87.89 & 53.85 & 66.78 & 80.16 & 17.78 & 29.10
& 65.63 & 62.17 \\

\bottomrule
\end{tabular*}
}
\caption{Label-level results on ClaimPlus. The labels in the dataset are Supported / Refuted / NEI / Conflicting Evidence.}
\label{tab:per-label-claimplus}
\end{table*}

\begin{table*}[t]
\centering
{\footnotesize
\renewcommand{\arraystretch}{1.10}
\setlength{\tabcolsep}{2pt}
\begin{tabular*}{\textwidth}{@{\extracolsep{\fill}}l*{14}{c}@{}}
\toprule
\multicolumn{1}{c}{\multirow{2}{*}{\textbf{Method}}}
& \multicolumn{3}{c}{\textbf{SUP}}
& \multicolumn{3}{c}{\textbf{REF}}
& \multicolumn{3}{c}{\textbf{NEI}}
& \multicolumn{3}{c}{\textbf{CON}}
& \multicolumn{2}{c}{\textbf{Overall}} \\
\cmidrule(lr){2-4}
\cmidrule(lr){5-7}
\cmidrule(lr){8-10}
\cmidrule(lr){11-13}
\cmidrule(l){14-15}
& \textbf{P} & \textbf{R} & \textbf{F1} & \textbf{P} & \textbf{R} & \textbf{F1} & \textbf{P} & \textbf{R} & \textbf{F1} & \textbf{P} & \textbf{R} & \textbf{F1} & \textbf{Acc} & \textbf{F1} \\
\midrule

\rowcolor{gray!12}
\multicolumn{15}{@{}c@{}}{
  \itshape\bfseries End-to-end Generation Models
} \\

DeepSeek-V4-Flash
& 51.27 & 78.41 & 62.00 & 32.38 & 51.03 & 39.62 & 43.89 & 11.16 & 17.80 & 17.16 & 7.24 & 10.18
& 45.99 & 32.40 \\

GPT-5.4
& 54.95 & 77.98 & 64.47 & 37.20 & 56.52 & 44.87 & 42.22 & 19.41 & 26.59 & 29.52 & 0.65 & 1.27
& 48.60 & 34.30 \\

Claude-Sonnet-4.6
& 53.47 & 81.50 & 64.57 & 33.95 & 50.59 & 40.63 & 58.02 & 8.23 & 14.42 & 9.31 & 5.84 & 7.18
& 46.19 & 31.70 \\

Qwen3-8B
& 61.94 & 57.82 & 59.81 & 34.01 & 64.50 & 44.54 & 56.05 & 21.44 & 31.02 & 11.73 & 20.17 & 14.83
& 43.91 & 37.55 \\

GPT-5.4 w/ Tools
& 55.50 & 73.87 & 63.38 & 32.27 & 73.19 & 44.79 & 36.80 & 6.25 & 10.69 & 9.38 & 1.34 & 2.34
& 45.60 & 30.30 \\

\rowcolor{gray!12}
\multicolumn{15}{@{}c@{}}{
  \itshape\bfseries OpenSearch Agents
} \\

MiroThink w/ Tools
& 48.74 & 67.84 & 56.73 & 34.59 & 44.91 & 39.08 & 30.99 & 33.76 & 32.32 & 15.46 & 12.13 & 13.59
& 47.95 & 35.43 \\

Qwen3-8B w/ Tools
& 61.37 & 60.21 & 60.78 & 37.74 & 62.00 & 46.92 & 48.82 & 10.34 & 17.07 & 12.86 & 32.03 & 18.35
& 42.28 & 35.78 \\

Tongyi-DR w/ Tools
& 55.72 & 70.65 & 62.30 & 33.90 & 56.62 & 42.41 & 30.43 & 6.51 & 10.73 & 9.80 & 14.83 & 11.80
& 42.93 & 31.81 \\

\rowcolor{gray!12}
\multicolumn{15}{@{}c@{}}{
  \itshape\bfseries AFC Systems
} \\

ClaimCheck
& 58.09 & 53.29 & 55.59 & 33.32 & 54.51 & 41.36 & 40.71 & 30.02 & 34.56 & 12.39 & 13.18 & 12.77
& 42.28 & 36.07 \\

DEFAME
& 60.83 & 43.59 & 50.79 & 28.61 & 60.32 & 38.81 & 37.76 & 30.62 & 33.82 & 11.67 & 11.19 & 11.42
& 39.09 & 33.71 \\

\rowcolor{gray!12}
\multicolumn{15}{@{}c@{}}{
  \itshape\bfseries Ours
} \\

OpenFC (Cold Start)
& 53.84 & 69.68 & 60.74 & 31.41 & 53.67 & 39.63 & 35.67 & 11.25 & 17.11 & 14.35 & 16.34 & 15.28
& 43.65 & 33.19 \\

OpenFC
& 72.40 & 71.71 & 72.05 & 44.72 & 82.61 & 58.03 & 82.70 & 5.70 & 10.66 & 23.51 & 12.34 & 16.18
& 47.17 & 39.23 \\

\bottomrule
\end{tabular*}
}
\caption{Label-level results on Climate-Fever. The labels in the dataset are Supported / Refuted / NEI / Conflicting Evidence.}
\label{tab:per-label-climate}
\end{table*}

\begin{table*}[t]
\centering
{\footnotesize
\renewcommand{\arraystretch}{1.10}
\setlength{\tabcolsep}{2pt}
\begin{tabular*}{\textwidth}{@{\extracolsep{\fill}}l*{11}{c}@{}}
\toprule
\multicolumn{1}{c}{\multirow{2}{*}{\textbf{Method}}}
& \multicolumn{3}{c}{\textbf{SUP}}
& \multicolumn{3}{c}{\textbf{REF}}
& \multicolumn{3}{c}{\textbf{NEI}}
& \multicolumn{2}{c}{\textbf{Overall}} \\
\cmidrule(lr){2-4}
\cmidrule(lr){5-7}
\cmidrule(lr){8-10}
\cmidrule(l){11-12}
& \textbf{P} & \textbf{R} & \textbf{F1} & \textbf{P} & \textbf{R} & \textbf{F1} & \textbf{P} & \textbf{R} & \textbf{F1} & \textbf{Acc} & \textbf{F1} \\
\midrule

\rowcolor{gray!12}
\multicolumn{12}{@{}c@{}}{
  \itshape\bfseries End-to-end Generation Models
} \\

DeepSeek-V4-Flash
& 45.63 & 77.57 & 57.46 & 40.96 & 33.81 & 37.04 & 79.77 & 31.92 & 45.60
& 44.53 & 46.70 \\

GPT-5.4
& 49.13 & 84.16 & 62.04 & 52.11 & 29.60 & 37.75 & 78.13 & 60.05 & 67.91
& 61.47 & 55.90 \\

Claude-Sonnet-4.6
& 36.04 & 83.26 & 50.30 & 38.16 & 31.25 & 34.36 & 33.08 & 3.26 & 5.94
& 29.47 & 30.20 \\

Qwen3-8B
& 52.84 & 59.90 & 56.15 & 33.98 & 28.00 & 30.70 & 78.77 & 34.98 & 48.45
& 40.53 & 45.10 \\

GPT-5.4 w/ Tools
& 52.31 & 80.20 & 63.32 & 42.80 & 49.60 & 45.95 & 78.67 & 46.10 & 58.13
& 55.87 & 55.80 \\

\rowcolor{gray!12}
\multicolumn{12}{@{}c@{}}{
  \itshape\bfseries OpenSearch Agents
} \\

MiroThink w/ Tools
& 56.44 & 48.23 & 52.01 & 45.12 & 21.93 & 29.51 & 71.61 & 46.38 & 56.30
& 42.80 & 45.94 \\

Qwen3-8B w/ Tools
& 46.20 & 66.92 & 54.66 & 41.65 & 22.45 & 29.17 & 76.01 & 14.60 & 24.50
& 30.00 & 36.11 \\

Tongyi-DR w/ Tools
& 30.02 & 63.32 & 40.73 & 25.27 & 28.21 & 26.66 & 37.99 & 23.84 & 29.30
& 35.20 & 32.23 \\

\rowcolor{gray!12}
\multicolumn{12}{@{}c@{}}{
  \itshape\bfseries AFC Systems
} \\

ClaimCheck
& 66.43 & 52.99 & 58.95 & 48.89 & 39.94 & 43.96 & 71.01 & 64.08 & 67.37
& 57.07 & 56.76 \\

DEFAME
& 70.62 & 51.56 & 59.60 & 39.67 & 38.88 & 39.27 & 72.64 & 62.95 & 67.45
& 55.87 & 55.44 \\

\rowcolor{gray!12}
\multicolumn{12}{@{}c@{}}{
  \itshape\bfseries Ours
} \\

OpenFC (Cold Start)
& 50.23 & 72.38 & 59.30 & 37.02 & 37.17 & 37.09 & 77.02 & 26.08 & 38.97
& 40.40 & 45.12 \\

OpenFC
& 56.33 & 90.00 & 69.29 & 36.38 & 63.99 & 46.39 & 83.25 & 39.76 & 53.82
& 57.33 & 56.50 \\

\bottomrule
\end{tabular*}
}
\caption{Label-level results on HealthFC. The labels in the dataset are Supported / Refuted / NEI.}
\label{tab:per-label-healthfc}
\end{table*}

\FloatBarrier

\newpage

\begin{table}[htbp]
\centering
\footnotesize
\renewcommand{\arraystretch}{1.0}
\setlength{\tabcolsep}{4pt}
\begin{tabular}{ll}
\toprule
\textbf{Configuration} & \textbf{Value} \\

\midrule
\multicolumn{2}{l}{\textit{RL algorithm}} \\
Algorithm                  & GRPO \\
Loss aggregation           & Token mean \\
Rollouts per prompt        & 8 \\
Advantage normalization    & Group mean / std \\
Dynamic sampling           & Enabled \\
Clip range                 & $[0.2, 0.3]$ \\
Dual-clip coefficient      & $3.0$ \\
KL loss                    & \texttt{low\_var\_kl} \\
KL coefficient             & $1 \times 10^{-3}$ \\
Reward KL penalty          & Disabled \\
Entropy coefficient        & $0$ \\
Rollout correction         & \texttt{seq\_mean\_k1} \\
Correction range           & $[0.1, 10.0]$ \\
\midrule
\multicolumn{2}{l}{\textit{Optimization}} \\
Optimizer                  & AdamW \\
Learning rate              & $2 \times 10^{-6}$ \\
Weight decay               & $0.01$ \\
Gradient clipping          & $1.0$ \\
Prompts per step           & 64 \\
PPO mini-batch             & 32 prompts \\
PPO epochs                 & 5 \\
Training steps             & 300 \\
\midrule
\multicolumn{2}{l}{\textit{Rollout}} \\
Inference engine           & SGLang \\
Temperature                & $1.0$ \\
Top-$p$                    & $1.0$ \\
Tool format                & Hermes \\
Tools                      & \texttt{search}, \texttt{visit} \\
Max assistant turns        & 15 \\
Max tool calls per turn    & 2 \\
Max tool response length   & 4{,}096 \\
\midrule
\multicolumn{2}{l}{\textit{Infrastructure}} \\
Hardware                   & $8 \times$ A100-80GB \\
Training parallelism       & FSDP, SP $=8$ \\
Rollout tensor parallelism & TP $=2$ \\
\bottomrule
\end{tabular}
\caption{Key reinforcement learning configuration.}
\label{tab:rl-config}
\end{table}

\section{Implementation Details}
\label{app:reward}

\subsection{RL training configuration}
Table~\ref{tab:rl-config} summarizes the GRPO training setup, including optimization, tool-augmented rollout, and distributed infrastructure configurations.
The 300-step run uses eight NVIDIA A100-80GB GPUs for 4.5 days wall-clock.

\begin{table}[tb]
\centering
\footnotesize
\renewcommand{\arraystretch}{1.08}
\setlength{\tabcolsep}{4pt}
\begin{tabular}{ll}
\toprule
\textbf{Configuration} & \textbf{Value} \\
\midrule
\multicolumn{2}{l}{\textit{Decoding}} \\
Inference engine       & vLLM \\
Temperature            & $0.85$ \\
Top-$p$                & $0.95$ \\
Presence penalty       & $1.1$ \\
Max tokens per turn    & 6{,}000 \\
Max model length       & 65{,}536 \\
\midrule
\multicolumn{2}{l}{\textit{Agent}} \\
Tools                  & \texttt{search}, \texttt{visit} \\
Max assistant turns    & 30 \\
Search backend         & Serper (Google) \\
Page summarizer        & \texttt{gpt-oss-20b} \\
Rollouts per claim     & 1 \\
Page fetch timeout         & 600 s \\

\bottomrule
\end{tabular}
\caption{Key inference configuration.}
\label{tab:infer-config}
\end{table}

\subsection{Inference configuration}

Table~\ref{tab:infer-config} summarizes the inference configuration used for all evaluations: every model is served with vLLM and queried through the same ReAct loop with \texttt{search} and \texttt{visit} tools, so that reported differences stem from the policy rather than the decoding or tool setup.


\section{Fact-Check Blocklist}
\label{app:blacklist}

We apply the same fact-check blocklist during training, inference, and
shared-interface evaluation. Its purpose is to prevent an agent from retrieving
an existing professional verdict instead of verifying a claim against
underlying evidence. A URL is blocked by either of the following two strategies.

\paragraph{Blocked fact-checking domains.}
We block dedicated fact-checking domains using exact hostname and subdomain
suffix matching. Thus, a listed domain and all of its subdomains are blocked,
whereas an unrelated hostname that merely contains the same string remains
accessible. For example, \url{politifact.com} and
\url{static.politifact.com} are blocked, but \url{notpolitifact.com} is not.
The blocked domains are:
\begin{itemize}
    \setlength{\itemsep}{0pt}
    \setlength{\parskip}{0pt}
    \setlength{\parsep}{0pt}
    \item \url{snopes.com}
    \item \url{politifact.com}
    \item \url{factcheck.org}
    \item \url{fullfact.org}
    \item \url{africacheck.org}
    \item \url{boomlive.in}
    \item \url{altnews.in}
    \item \url{factly.in}
    \item \url{leadstories.com}
    \item \url{checkyourfact.com}
    \item \url{vishvasnews.com}
    \item \url{newschecker.in}
    \item \url{factcrescendo.com}
    \item \url{verafiles.org}
    \item \url{newsmeter.in}
    \item \url{factcheckhub.com}
    \item \url{mediabiasfactcheck.com}
    \item \url{logically.ai}
    \item \url{logicallyfacts.com}
    \item \url{poynter.org}
    \item \url{factcheck.afp.com}
    \item \url{fastcheck.cl}
    \item \url{teyit.org}
    \item \url{maldita.es}
    \item \url{newtral.es}
    \item \url{factcheck.kz}
    \item \url{dubawa.org}
    \item \url{factcheck.ge}
    \item \url{factcheckni.org}
\end{itemize}
The entry \url{factcheck.afp.com} blocks only AFP's fact-checking subdomain;
the general news domain \url{afp.com} remains accessible.

\paragraph{Blocked fact-checking URL patterns.}
General-purpose news domains are not blocked in their entirety. Instead, we
lowercase the complete URL and block it only when it contains one of the
following fact-checking path patterns:
\begin{itemize}
    \setlength{\itemsep}{0pt}
    \setlength{\parskip}{0pt}
    \setlength{\parsep}{0pt}
    \item \texttt{/fact-check}
    \item \texttt{/factcheck}
    \item \texttt{/fact\_check}
    \item \texttt{/fact-checking}
    \item \texttt{/webqoof}
    \item \texttt{ap-fact-check}
    \item \texttt{/factsfirst}
    \item \texttt{/news/fact-checker}
    \item \texttt{/hub/ap-fact-check}
\end{itemize}
This case-insensitive path matching blocks dedicated fact-checking sections
while retaining access to ordinary reporting from the same publisher. For
example, \texttt{/webqoof} covers The Quint's fact-checking section without
blocking the remainder of its website.

\section{Prompt and Supervisor Decision Contracts}
\label{app:prompt}

This appendix specifies the interaction contracts used to construct the
stepwise-calibrated trajectories.
The verification policy interacts with the open-web environment through
\textsc{Search} and \textsc{Visit}.
Retrieved webpages are processed by a goal-conditioned evidence summarizer
before being returned as observations.
Before an action is executed, the trajectory supervisor examines the proposed
transition using the claim, cutoff timestamp, immediately preceding policy
step, latest environment observation, and current round index.

The supervisor is never given the reference verdict.
It returns \textsc{Keep} when the proposed transition is evidence-grounded,
well formed, and useful, and \textsc{Modify} otherwise.
An accepted tool action is executed before the next proposal, making its real
environment response part of the subsequent supervision context.
When a terminal proposal must be modified, the supervisor supplies only a
revised reasoning continuation and requests terminal regeneration by the
rollout policy; it does not directly insert a replacement verdict.

\subsection{Verification Prompt}
\label{app:verification-prompt}

The rollout policy receives the following system prompt throughout the
trajectory.

\begin{lstlisting}[style=openfcprompt,title={\bfseries Verification Policy: System Prompt}]
You are an open-web fact-checking agent. Your task is to investigate a claim using external evidence and determine whether the available evidence supports, refutes, fails to resolve, or materially conflicts about the claim.

You must actively use the provided tools before producing a final judgment. Base your reasoning only on the claim, the evidence returned by the environment, and information available on or before the specified evidence cutoff timestamp. Do not invent observations, webpage contents, sources, URLs, or tool results.

Search for evidence that could both support and contradict the claim. Prefer credible, relevant, and clearly dated sources. Focus on resolving the central factual assertion rather than collecting large amounts of tangential information. Stop searching once the available evidence is sufficient for a justified decision.

At each round, return exactly one of the following forms.

Non-terminal response:

<think>
A concise evidence-grounded explanation of the next information need.
</think>
<tool_call>
{"name": "<function-name>", "arguments": <arguments-json-object>}
</tool_call>

Terminal response:

<think>
A concise synthesis of the evidence supporting the decision.
</think>
<answer>
## Conclusion <Label>
Label: <Label>
Reason: <brief claim-focused justification>
</answer>

Do not return a tool call and a final answer in the same response. Do not generate or simulate a tool response yourself.

# Tools

You are provided with the following function signatures:

<tools>
{"type": "function", "function": {"name": "search", "description": "Perform open-web searches and return the top search results for one or more queries.", "parameters": {"type": "object", "properties": {"query": {"type": "array", "items": {"type": "string", "description": "A claim-specific search query."}, "minItems": 1, "description": "A small set of diverse, non-redundant search queries."}}, "required": ["query"]}}}

{"type": "function", "function": {"name": "visit", "description": "Visit one or more webpages returned by the search environment and extract information relevant to a specified goal.", "parameters": {"type": "object", "properties": {"url": {"type": "array", "items": {"type": "string"}, "minItems": 1, "description": "URLs that have already appeared in a previous tool response."}, "goal": {"type": "string", "description": "The specific evidentiary question to resolve from the webpages."}}, "required": ["url", "goal"]}}} </tools>

Return each function call as one valid JSON object enclosed in <tool_call></tool_call> tags:

<tool_call>
{"name": "<function-name>", "arguments": <arguments-json-object>}
</tool_call>
\end{lstlisting}

Each instance is then initialized with the following task prompt.

\begin{lstlisting}[style=openfcprompt,title={\bfseries Verification Policy: Task Prompt}]
Verify the following claim through open-domain search. You must retrieve and examine external evidence before making a judgment.

Use exactly one of the following labels:

* Supported: The available evidence directly and sufficiently confirms the core content of the claim. Minor wording differences are allowed, but the main factual assertion is upheld.

* Refuted: The best available evidence directly contradicts the core content of the claim or establishes that the claim is materially false in a key respect.

* NEI: The available evidence is insufficient to verify or falsify the claim. This includes cases where the claim is too vague, reliable evidence cannot be found, or the available evidence is too indirect for a firm conclusion.

* Conflicting Evidence: Credible evidence materially disagrees about the core content of the claim, with some evidence supporting it and other evidence contradicting it.

Claim: {claim}

Evidence Cutoff Timestamp: {cutoff_timestamp}

The final answer must use the following format:

<answer>
## Conclusion <Label>
Label: <Supported / Refuted / NEI / Conflicting Evidence>
Reason: <brief claim-focused justification>
</answer>

The label in the heading and the Label field must be identical.
\end{lstlisting}

\subsection{Webpage Evidence Summarization Prompt}
\label{app:summary-prompt}

The \textsc{Visit} environment uses a goal-conditioned summarizer to convert
raw webpage content into an evidence-focused observation.
The summarizer is instructed to preserve the source's dates, quantities,
entities, qualifications, and surrounding context while avoiding unsupported
inferences.

\begin{lstlisting}[style=openfcprompt,title={\bfseries Goal-Conditioned Webpage Summarization Prompt}]
You are an evidence-extraction module for an open-web fact-checking system.

Given webpage content and a specific information goal, identify the portions of the page that are directly relevant to that goal. Do not introduce facts, interpretations, or source claims that are absent from the provided content.

# Webpage Content

{webpage_content}

# Information Goal

{goal}

# Extraction Requirements

1. Locate the specific passages, statements, tables, dates, quantities, and named entities that address the information goal.

2. Preserve the relevant context needed to interpret each piece of evidence correctly, including qualifications, uncertainty, temporal scope, and whether the source is reporting, quoting, or making the claim itself.

3. Retain exact dates, numerical values, entity names, and source attributions whenever they are relevant.

4. Distinguish direct evidence from indirect or contextual information. Do not treat the absence of information on the page as evidence that an event did not occur.

5. If the page contains materially inconsistent statements, preserve both sides of the inconsistency.

6. If no relevant evidence is present, state this explicitly rather than constructing an answer from unrelated content.

Return one valid JSON object and nothing else. Do not wrap the JSON in Markdown fences.

{
"rationale": "Why the extracted content is relevant, partially relevant, or not relevant to the information goal.",
"evidence": "The relevant source content with sufficient original context, preserving important dates, quantities, entities, qualifications, and attributions.",
"summary": "A concise synthesis of what the webpage contributes to resolving the information goal."
}
\end{lstlisting}

\subsection{Supervisor Prompt}
\label{app:supervisor-prompt}

The trajectory supervisor evaluates the rollout policy's proposed transition
before it changes the evidence state.
A value of \texttt{false} for \texttt{was\_modified} corresponds to
\textsc{Keep}, whereas \texttt{true} corresponds to \textsc{Modify}.

\begin{lstlisting}[style=openfcprompt,title={\bfseries Trajectory Supervisor: Calibration Prompt}]
You are a fact-checking trajectory calibration expert. Review the fact-checking model's current reasoning and proposed action before that action affects the subsequent evidence trajectory.

You will receive:

1. The claim and evidence cutoff timestamp.
2. The immediately preceding model step, if available.
3. The latest tool response, if available.
4. The model's current proposed response.
5. The current round index.

The current response will contain one of the following:

* <think>...</think> followed by <tool_call>...</tool_call>;
* <think>...</think> followed by <answer>...</answer>; or
* a standalone <answer>...</answer>.

Your task is to determine whether the proposed transition should be kept or modified.

# Calibration Criteria

Check whether:

* The reasoning is logically sound and follows from the claim, immediately preceding model step, and latest tool response.

* The reasoning does not invent observations, sources, URLs, webpage contents, or facts that have not been returned by the environment.

* The proposed action targets the most important unresolved part of the claim.

* The proposed tool call is syntactically valid, necessary, effective, and reasonably efficient.

* A search query is claim-specific and likely to resolve a material uncertainty.

* A visit action targets previously returned URLs whose contents are likely to provide useful or decisive evidence.

* A terminal decision is sufficiently supported by the evidence already obtained.

* The response respects the evidence cutoff timestamp.

Do not use, infer, or reconstruct a hidden reference verdict. Calibrate only from the claim, interaction context, and evidence returned by the tools.

# Evidence Cutoff

Do not rely on information that became available after the evidence cutoff timestamp.

If a tool response contains post-cutoff information, ignore that information when assessing the claim. Prefer contemporaneous or clearly dated evidence that was publicly available on or before the cutoff.

Pages published after the cutoff or without a resolvable publication date are filtered by the environment and must not be used.

# Tool Constraints

Only two tools exist:

* search
* visit

Preserve the existing tool-call syntax exactly. Do not invent a new function, argument, or schema.

For search:

* Use a small number of claim-specific queries.
* Make the queries diverse and non-overlapping.
* Remove redundant, overly broad, or speculative queries.
* Prefer queries that directly resolve the central factual uncertainty.

For visit:

* Every visited URL must have appeared in an earlier tool response.
* Do not fabricate, complete, or modify a URL.
* Visit a page only when its content is likely to contribute material evidence.
* Do not repeatedly visit a page that has already failed or returned unusable content; switch source or strategy instead.

Use at most one tool action in the calibrated response.

# Trajectory Control

* Keep the trajectory within 35 rounds.
* Focus on the central unresolved part of the claim.
* Avoid broad, repetitive, or drifting exploration.
* Prefer decisive evidence and synthesis over speculative retrieval.
* If the evidence is insufficient, select the single action with the highest expected evidentiary value.
* If the evidence is already sufficient, do not preserve an unnecessary tool action.
* If Round >= 30, allow at most one further high-value tool action; otherwise move toward termination.
* If Round >= 35, do not issue another tool call.

# Output States

Return exactly one of the following states.

## 1. Tool-Action State

Use this state when another environment action remains necessary.

* calibrated_thinking contains exactly one concise <think>...</think> block.
* calibrated_tool_call contains exactly one <tool_call>...</tool_call> block.
* calibrated_answer is an empty string.

This state may either preserve a valid proposed action or replace it with a better action.

## 2. Accepted-Terminal State

Use this state only when the model's current terminal answer is already sufficiently supported and its substantive label and justification can be retained.

* calibrated_thinking contains exactly one concise <think>...</think> block.
* calibrated_answer contains exactly one <answer>...</answer> block.
* calibrated_tool_call is an empty string.

Do not use this state to introduce a new verdict that was absent from, or substantively different from, the model's current terminal proposal.

## 3. Regeneration State

Use this state when:

* an unsupported or malformed terminal proposal must be revised;
* a proposed tool action should instead terminate because the existing evidence is sufficient; or
* the round limit requires termination but the current response is not an acceptable terminal answer.

In this state:

* calibrated_thinking contains exactly one concise <think>...</think> block that corrects the reasoning direction and indicates that a terminal answer should now be generated;
* calibrated_answer is an empty string;
* calibrated_tool_call is an empty string.

Do not directly supply a replacement verdict in the regeneration state. The rollout model will generate a new terminal answer from the calibrated continuation.

# Terminal-Answer Requirements

An accepted terminal answer must:

* directly resolve the original claim;
* use exactly one of the following labels:

  * Supported
  * Refuted
  * NEI
  * Conflicting Evidence
* contain only the minimum evidence needed to justify the decision;
* exclude search narration, procedural details, unsupported claims, and tangential information.

The exact answer format is:

<answer>
## Conclusion <Label>
Label: <Supported / Refuted / NEI / Conflicting Evidence>
Reason: <brief claim-focused justification>
</answer>

The label in the heading and the Label field must be identical.

# Modification Decision

Set was_modified to false only when the substantive reasoning, action, or accepted terminal answer is preserved.

Set was_modified to true if you:

* revise the reasoning;
* change, remove, or replace a tool action;
* redirect the trajectory toward termination;
* reject or require regeneration of a terminal answer; or
* make any other substantive correction.

A purely superficial formatting normalization may preserve was_modified as false only when it does not change the reasoning, action, verdict, or justification.

calibration_reasoning must briefly state why the response was kept or modified. It must not contain hidden-label information, extended chain-of-thought, or a separate fact-checking conclusion.

# Output Format

Return one valid JSON object and nothing else.

Always include all five keys. Use an empty string for every inapplicable field.

{
"was_modified": true,
"calibration_reasoning": "Brief explanation of why the proposed transition was kept or modified.",
"calibrated_thinking": "<think>...</think>",
"calibrated_answer": "",
"calibrated_tool_call": "<tool_call>...</tool_call>"
}

Inputs:

Claim: {claim}

Evidence Cutoff Timestamp: {cutoff_timestamp}

Round: {round_idx}

Immediately Preceding Model Step:
{previous_model_step}

Latest Tool Response:
{last_tool_response}

Model's Current Proposed Response:
{full_response}
\end{lstlisting}

The supervisor output is consumed before environment execution.
In the tool-action state, the calibrated action is executed and its observation
is appended to the trajectory.
In the accepted-terminal state, the retained answer terminates the rollout.
In the regeneration state, the calibrated reasoning continuation is returned
to the rollout policy, which generates a new terminal answer without receiving
the reference verdict.

\fi


\end{document}